\pdfoutput=1
\documentclass{article} 
\usepackage{iclr2027_conference,times}

\usepackage{amsmath,amsfonts,bm}

\def\eqref#1{equation~\ref{#1}}

\def\1{\bm{1}}

\DeclareMathAlphabet{\mathsfit}{\encodingdefault}{\sfdefault}{m}{sl}
\SetMathAlphabet{\mathsfit}{bold}{\encodingdefault}{\sfdefault}{bx}{n}

\usepackage{booktabs}       
\usepackage{wrapfig}        
\usepackage{hyperref}
\ifdefined\pdfrunninglinkoff
\makeatletter
\protected\def\pdfannot@link@off@@{\pdfrunninglinkoff}
\protected\def\pdfannot@link@on@@{\pdfrunninglinkon}
\AddToHook{cmd/@makecol/before}{%
  \ifvoid\footins\else
    \setbox\footins\vbox{\pdfrunninglinkoff\unvbox\footins\pdfrunninglinkon}%
  \fi}
\makeatother
\fi
\usepackage{url}

\PassOptionsToPackage{dvipsnames,table}{xcolor}
\usepackage{xcolor}
\usepackage{tikz}
\usetikzlibrary{arrows.meta, positioning, calc,fit,backgrounds}
\pgfdeclarelayer{midbg}
\pgfdeclarelayer{foreground}
\pgfsetlayers{background,midbg,main,foreground}
\usepackage{pgfplots}
\usepgfplotslibrary{statistics}   
\usepackage{pgfplotstable}        
\usetikzlibrary{arrows.meta,decorations.text}
\usepgfplotslibrary{groupplots}
\usepackage{pifont}
\usepackage{bbm} 
\usepackage{amssymb}

\usepackage{siunitx}
\usepackage{multirow}
\usepackage{makecell}
\usepackage{enumitem}
\usepackage{tcolorbox}
\tcbuselibrary{listings, breakable}
\usepackage{arydshln}
\usepackage{etoc} 
\newcounter{promptctr}
\usepackage{subcaption}
\usepackage[table]{xcolor}

\usepackage{colortbl}

\definecolor{brandblue}{rgb}{0.34, 0.7, 1}
\tcbset {
	base/.style={
		arc=0mm, 
		bottomtitle=0.5mm,
		boxrule=0mm,
		colbacktitle=black!10!white, 
		coltitle=black, 
		fonttitle=\bfseries, 
		left=2.5mm,
		leftrule=1mm,
		right=3.5mm,
		title={#1},
		toptitle=0.75mm, 
	}
}
\newtcolorbox{promptbox}[2][]{%
	breakable,
	colback=gray!5!white,
	colframe=brandblue,
	base={#1},
	before upper={%
		\refstepcounter{promptctr}%
		\ifx#1\empty\else\label{#1}\fi
	},
	title={Prompt~\number\numexpr\value{promptctr}+1\relax: #2},
}

\usepackage[frozencache=true,cachedir=minted-cache]{minted}
\tcbuselibrary{minted}
\definecolor{xmlcolor}{HTML}{225496}
\newtcblisting{xmloutputbox}{
	listing only,
	listing engine=minted,
	minted language=xml,
	listing options={tabsize=2},
	minted options={
		tabsize=0,
		fontsize=\footnotesize,
		style=tango
	},
	frame empty,
	left=0pt,
	right=0pt,
	top=0pt,
	bottom=0pt,
	boxsep=0pt,
	boxrule=0pt,
	colback=gray!5!white
}
\newtcblisting{xmlbox}{
	listing engine=minted,
	minted language=xml,
	minted style=tango,
	minted options={
		fontsize=\tiny,
		breaklines,
		tabsize=2
	},
	colback=white,
	colframe=black,
	boxrule=0.4pt,
	arc=2pt,
	left=4pt,
	right=4pt,
	top=3pt,
	bottom=3pt
}
\newtcblisting{xmlcodebox}{
	listing only,
	listing engine=minted,
	minted language=xml,
	listing options={tabsize=2},
	minted options={
		tabsize=2,
		fontsize=\footnotesize,
		style=tango,
		    breaklines=true,
		breakafter={/}
	},
	left=3pt,
	right=3pt,
	top=3pt,
	bottom=3pt,
	boxsep=0pt,
	boxrule=0pt,
	colback=blue!5
}
\newtcolorbox{structmetabox}{
	colframe=red!70!black,
	colback=white!0,
	boxrule=0.6pt,
	left=6pt,
	right=6pt,
	top=4pt,
	bottom=4pt,
}

\makeatletter
\def\PYG@reset{\let\PYG@it=\relax \let\PYG@bf=\relax%
	\let\PYG@ul=\relax \let\PYG@tc=\relax%
	\let\PYG@bc=\relax \let\PYG@ff=\relax}
\def\PYG@tok#1{\csname PYG@tok@#1\endcsname}
\def\PYG@toks#1+{\ifx\relax#1\empty\else%
	\PYG@tok{#1}\expandafter\PYG@toks\fi}
\def\PYG@do#1{\PYG@bc{\PYG@tc{\PYG@ul{%
				\PYG@it{\PYG@bf{\PYG@ff{#1}}}}}}}
\def\PYG#1#2{\PYG@reset\PYG@toks#1+\relax+\PYG@do{#2}}

\@namedef{PYG@tok@w}{\def\PYG@tc##1{\textcolor[rgb]{0.73,0.73,0.73}{##1}}}
\@namedef{PYG@tok@c}{\let\PYG@it=\textit\def\PYG@tc##1{\textcolor[rgb]{0.00,0.60,1.00}{##1}}}
\@namedef{PYG@tok@cp}{\def\PYG@tc##1{\textcolor[rgb]{0.00,0.60,0.60}{##1}}}
\@namedef{PYG@tok@cs}{\let\PYG@bf=\textbf\let\PYG@it=\textit\def\PYG@tc##1{\textcolor[rgb]{0.00,0.60,1.00}{##1}}}
\@namedef{PYG@tok@k}{\let\PYG@bf=\textbf\def\PYG@tc##1{\textcolor[rgb]{0.00,0.40,0.60}{##1}}}
\@namedef{PYG@tok@kp}{\def\PYG@tc##1{\textcolor[rgb]{0.00,0.40,0.60}{##1}}}
\@namedef{PYG@tok@kt}{\let\PYG@bf=\textbf\def\PYG@tc##1{\textcolor[rgb]{0.00,0.47,0.53}{##1}}}
\@namedef{PYG@tok@o}{\def\PYG@tc##1{\textcolor[rgb]{0.33,0.33,0.33}{##1}}}
\@namedef{PYG@tok@ow}{\let\PYG@bf=\textbf\def\PYG@tc##1{\textcolor[rgb]{0.00,0.00,0.00}{##1}}}
\@namedef{PYG@tok@nb}{\def\PYG@tc##1{\textcolor[rgb]{0.20,0.40,0.40}{##1}}}
\@namedef{PYG@tok@nf}{\def\PYG@tc##1{\textcolor[rgb]{0.80,0.00,1.00}{##1}}}
\@namedef{PYG@tok@nc}{\let\PYG@bf=\textbf\def\PYG@tc##1{\textcolor[rgb]{0.00,0.67,0.53}{##1}}}
\@namedef{PYG@tok@nn}{\let\PYG@bf=\textbf\def\PYG@tc##1{\textcolor[rgb]{0.00,0.80,1.00}{##1}}}
\@namedef{PYG@tok@ne}{\let\PYG@bf=\textbf\def\PYG@tc##1{\textcolor[rgb]{0.80,0.00,0.00}{##1}}}
\@namedef{PYG@tok@nv}{\def\PYG@tc##1{\textcolor[rgb]{0.00,0.20,0.20}{##1}}}
\@namedef{PYG@tok@no}{\def\PYG@tc##1{\textcolor[rgb]{0.20,0.40,0.00}{##1}}}
\@namedef{PYG@tok@nl}{\def\PYG@tc##1{\textcolor[rgb]{0.60,0.60,1.00}{##1}}}
\@namedef{PYG@tok@ni}{\let\PYG@bf=\textbf\def\PYG@tc##1{\textcolor[rgb]{0.60,0.60,0.60}{##1}}}
\@namedef{PYG@tok@na}{\def\PYG@tc##1{\textcolor[rgb]{0.20,0.00,0.60}{##1}}}
\@namedef{PYG@tok@nt}{\let\PYG@bf=\textbf\def\PYG@tc##1{\textcolor[rgb]{0.20,0.00,0.60}{##1}}}
\@namedef{PYG@tok@nd}{\def\PYG@tc##1{\textcolor[rgb]{0.60,0.60,1.00}{##1}}}
\@namedef{PYG@tok@s}{\def\PYG@tc##1{\textcolor[rgb]{0.80,0.20,0.00}{##1}}}
\@namedef{PYG@tok@sd}{\let\PYG@it=\textit\def\PYG@tc##1{\textcolor[rgb]{0.80,0.20,0.00}{##1}}}
\@namedef{PYG@tok@si}{\def\PYG@tc##1{\textcolor[rgb]{0.67,0.00,0.00}{##1}}}
\@namedef{PYG@tok@se}{\let\PYG@bf=\textbf\def\PYG@tc##1{\textcolor[rgb]{0.80,0.20,0.00}{##1}}}
\@namedef{PYG@tok@sr}{\def\PYG@tc##1{\textcolor[rgb]{0.20,0.67,0.67}{##1}}}
\@namedef{PYG@tok@ss}{\def\PYG@tc##1{\textcolor[rgb]{1.00,0.80,0.20}{##1}}}
\@namedef{PYG@tok@sx}{\def\PYG@tc##1{\textcolor[rgb]{0.80,0.20,0.00}{##1}}}
\@namedef{PYG@tok@m}{\def\PYG@tc##1{\textcolor[rgb]{1.00,0.40,0.00}{##1}}}
\@namedef{PYG@tok@gh}{\let\PYG@bf=\textbf\def\PYG@tc##1{\textcolor[rgb]{0.00,0.20,0.00}{##1}}}
\@namedef{PYG@tok@gu}{\let\PYG@bf=\textbf\def\PYG@tc##1{\textcolor[rgb]{0.00,0.20,0.00}{##1}}}
\@namedef{PYG@tok@gd}{\def\PYG@bc##1{{\setlength{\fboxsep}{\string -\fboxrule}\fcolorbox[rgb]{0.80,0.00,0.00}{1.00,0.80,0.80}{\strut ##1}}}}
\@namedef{PYG@tok@gi}{\def\PYG@bc##1{{\setlength{\fboxsep}{\string -\fboxrule}\fcolorbox[rgb]{0.00,0.80,0.00}{0.80,1.00,0.80}{\strut ##1}}}}
\@namedef{PYG@tok@gr}{\def\PYG@tc##1{\textcolor[rgb]{1.00,0.00,0.00}{##1}}}
\@namedef{PYG@tok@ge}{\let\PYG@it=\textit}
\@namedef{PYG@tok@gs}{\let\PYG@bf=\textbf}
\@namedef{PYG@tok@ges}{\let\PYG@bf=\textbf\let\PYG@it=\textit}
\@namedef{PYG@tok@gp}{\let\PYG@bf=\textbf\def\PYG@tc##1{\textcolor[rgb]{0.00,0.00,0.60}{##1}}}
\@namedef{PYG@tok@go}{\def\PYG@tc##1{\textcolor[rgb]{0.67,0.67,0.67}{##1}}}
\@namedef{PYG@tok@gt}{\def\PYG@tc##1{\textcolor[rgb]{0.60,0.80,0.40}{##1}}}
\@namedef{PYG@tok@err}{\def\PYG@tc##1{\textcolor[rgb]{0.67,0.00,0.00}{##1}}\def\PYG@bc##1{{\setlength{\fboxsep}{0pt}\colorbox[rgb]{1.00,0.67,0.67}{\strut ##1}}}}
\@namedef{PYG@tok@kc}{\let\PYG@bf=\textbf\def\PYG@tc##1{\textcolor[rgb]{0.00,0.40,0.60}{##1}}}
\@namedef{PYG@tok@kd}{\let\PYG@bf=\textbf\def\PYG@tc##1{\textcolor[rgb]{0.00,0.40,0.60}{##1}}}
\@namedef{PYG@tok@kn}{\let\PYG@bf=\textbf\def\PYG@tc##1{\textcolor[rgb]{0.00,0.40,0.60}{##1}}}
\@namedef{PYG@tok@kr}{\let\PYG@bf=\textbf\def\PYG@tc##1{\textcolor[rgb]{0.00,0.40,0.60}{##1}}}
\@namedef{PYG@tok@bp}{\def\PYG@tc##1{\textcolor[rgb]{0.20,0.40,0.40}{##1}}}
\@namedef{PYG@tok@fm}{\def\PYG@tc##1{\textcolor[rgb]{0.80,0.00,1.00}{##1}}}
\@namedef{PYG@tok@vc}{\def\PYG@tc##1{\textcolor[rgb]{0.00,0.20,0.20}{##1}}}
\@namedef{PYG@tok@vg}{\def\PYG@tc##1{\textcolor[rgb]{0.00,0.20,0.20}{##1}}}
\@namedef{PYG@tok@vi}{\def\PYG@tc##1{\textcolor[rgb]{0.00,0.20,0.20}{##1}}}
\@namedef{PYG@tok@vm}{\def\PYG@tc##1{\textcolor[rgb]{0.00,0.20,0.20}{##1}}}
\@namedef{PYG@tok@sa}{\def\PYG@tc##1{\textcolor[rgb]{0.80,0.20,0.00}{##1}}}
\@namedef{PYG@tok@sb}{\def\PYG@tc##1{\textcolor[rgb]{0.80,0.20,0.00}{##1}}}
\@namedef{PYG@tok@sc}{\def\PYG@tc##1{\textcolor[rgb]{0.80,0.20,0.00}{##1}}}
\@namedef{PYG@tok@dl}{\def\PYG@tc##1{\textcolor[rgb]{0.80,0.20,0.00}{##1}}}
\@namedef{PYG@tok@s2}{\def\PYG@tc##1{\textcolor[rgb]{0.80,0.20,0.00}{##1}}}
\@namedef{PYG@tok@sh}{\def\PYG@tc##1{\textcolor[rgb]{0.80,0.20,0.00}{##1}}}
\@namedef{PYG@tok@s1}{\def\PYG@tc##1{\textcolor[rgb]{0.80,0.20,0.00}{##1}}}
\@namedef{PYG@tok@mb}{\def\PYG@tc##1{\textcolor[rgb]{1.00,0.40,0.00}{##1}}}
\@namedef{PYG@tok@mf}{\def\PYG@tc##1{\textcolor[rgb]{1.00,0.40,0.00}{##1}}}
\@namedef{PYG@tok@mh}{\def\PYG@tc##1{\textcolor[rgb]{1.00,0.40,0.00}{##1}}}
\@namedef{PYG@tok@mi}{\def\PYG@tc##1{\textcolor[rgb]{1.00,0.40,0.00}{##1}}}
\@namedef{PYG@tok@il}{\def\PYG@tc##1{\textcolor[rgb]{1.00,0.40,0.00}{##1}}}
\@namedef{PYG@tok@mo}{\def\PYG@tc##1{\textcolor[rgb]{1.00,0.40,0.00}{##1}}}
\@namedef{PYG@tok@ch}{\let\PYG@it=\textit\def\PYG@tc##1{\textcolor[rgb]{0.00,0.60,1.00}{##1}}}
\@namedef{PYG@tok@cm}{\let\PYG@it=\textit\def\PYG@tc##1{\textcolor[rgb]{0.00,0.60,1.00}{##1}}}
\@namedef{PYG@tok@cpf}{\let\PYG@it=\textit\def\PYG@tc##1{\textcolor[rgb]{0.00,0.60,1.00}{##1}}}
\@namedef{PYG@tok@c1}{\let\PYG@it=\textit\def\PYG@tc##1{\textcolor[rgb]{0.00,0.60,1.00}{##1}}}

\makeatother

\title{MolLangData: A Large-Scale Dataset for \\Molecular Structure--Language Description via a Rule-Regularized Method}

\author{%
\hspace*{-\tabcolsep}\parbox[t]{\textwidth}{\raggedright
\normalsize\bfseries
Feiyang Cai\textsuperscript{1,*}, Guijuan He\textsuperscript{1}, Yi Hu\textsuperscript{2}, Jingjing Wang\textsuperscript{1}, Joshua Luo\textsuperscript{3},\\
Tianyu Zhu\textsuperscript{1}, Srikanth Pilla\textsuperscript{4}, Gang Li\textsuperscript{1}, Ling Liu\textsuperscript{3}, Feng Luo\textsuperscript{1,*}\\[0.5em]
\normalfont\small
\textsuperscript{1}Clemson University\enspace \textsuperscript{2}Independent Researcher\enspace \textsuperscript{3}Georgia Institute of Technology\enspace \textsuperscript{4}University of Delaware\\[0.4em]
\small \textsuperscript{*}Corresponding authors: \texttt{\{feiyang,luofeng\}@clemson.edu}
}\hspace*{-\tabcolsep}%
}

\iclrfinalcopy 

\begin{document}

\maketitle
\lhead{Preprint}
\begingroup
\renewcommand{\thefootnote}{}
\footnotetext{The dataset is hosted at \url{https://huggingface.co/datasets/ChemFM/MolLangData}, and the project code is available at \url{https://github.com/TheLuoFengLab/MolLangData}. For further details, see the \hyperref[sec:reproducibility]{Reproducibility Statement}.}
\endgroup
\begin{abstract}
	Molecular function is largely determined by structure.
	Accurately aligning molecular structure with natural language is therefore essential for enabling large language models (LLMs) to both understand   structure for chemical reasoning and generate molecules from natural-language design intent.
	However, the substantial cost of human annotation makes it infeasible to construct large-scale, high-quality datasets of structure-grounded descriptions.
	%
	%
	This work proposes a fully automated annotation framework for generating precise molecular descriptions at scale, such that the original molecule can be unambiguously reconstructed from the description alone.
	Our approach extends a rule-based chemical nomenclature parser to interpret IUPAC names and construct enriched,  XML metadata that explicitly encodes molecular structure. 
	This is then used to guide LLMs in producing accurate natural-language descriptions.
	Using this framework, we curate MolLangData, a dataset of approximately $163$k molecule--description pairs.
	A rigorous validation protocol combining expert human and LLM-based reconstruction on a subset of $\num{2000}$ molecules demonstrates $98.6\%$ description precision.
	Using the curated dataset, we train a $4$B-parameter LLM via large-scale reinforcement learning for language-conditional molecule generation.
	%
	Despite its much smaller size, the resulting model significantly outperforms most of the frontier LLMs while using far fewer tokens.
	The proposed framework and dataset provide a reliable foundation for molecule--language alignment, readily beneficial to broader chemical tasks.

	%

	%


\end{abstract}

\section{Introduction}
By aligning other modalities with language, the strong reasoning capabilities, flexibility, and extensive knowledge embedded in large language models (LLMs) can be transferred to multimodal settings, allowing for interpretation and inference beyond text.
This paradigm has proven particularly successful in vision--language models (VLMs).
For example, VLMs are now widely used for image recognition and analysis~\citep{openai5_6}, as well as for conditional image generation and editing based on user instructions~\citep{openaiimage2}.
Similarly, aligning molecular representations with language could enable chemical reasoning that is currently not possible for traditional unimodal models~\citep{chemfm,chemformer},
thereby providing new opportunities for diverse chemical tasks such as property prediction~\citep{moleculenet}, molecular and drug discovery~\citep{guacamol,moses}, and synthesis prediction and planning~\citep{usptofull}.

Early attempts to bridge molecular representations with language include the curation of molecule-description datasets~\citep{text2mol,L+M} and subsequent models that translate between the two modalities~\citep{text2mol,molt5}.
With the emergence of LLMs, research efforts have shifted toward leveraging them for downstream chemical tasks, including direct capability benchmarking~\citep{chemllmbench}, training chemistry- or science-specific LLMs~\citep{galactica,chemllm}, and building multimodal models~\citep{momu,jablonka2024leveraging,llamole}.
However, these approaches remain far from practical use and still lag behind unimodal chemical models on nearly all downstream tasks.
%
More detailed related work is discussed in Appendix~\ref{sec:related_works}.

\begin{figure}[t]
	\centering
		\resizebox{1\textwidth}{!}{
			\input{figures/intro/main_figure.tex}
			}
	\caption{An illustrative example motivating this work. Existing approaches align molecular representations with high-level objectives; we argue molecule--language alignment should be structure-grounded, with higher-level reasoning left to the LLM backbone, analogous to image--language alignment. 
	Real molecular descriptions are far more complex; see Appendix Fig.~\ref{fig:running_example_descriptions} and Sec.~\ref{sec:data_samples}.
	}
	\label{fig:motivation_illustration}
	\vspace{-0.25in}
\end{figure}

These models, along with curated datasets, bypass alignment between language and the underlying molecular structure. 
Instead, they attempt to directly bridge the molecular representations with downstream objectives that require higher-level, domain-specific reasoning.
However, as illustrated in Fig.~\ref{fig:motivation_illustration}, a molecule's function, including physicochemical properties and reaction mechanisms,  is determined by its structure.
Whether performing property inference or molecular design,  chemists always rely on an explicit understanding of molecular structure, and so should AI models.
This gap has been systematically highlighted by MolLangBench~\citep{mollangbench}, which evaluates molecule--language interface tasks such as molecular structure recognition, and structure generation and editing from language. 
The results show that even advanced general-purpose LLMs perform far from perfectly on these tasks, while chemical-specific multimodal models fail almost entirely.
Without these foundational capabilities, success on more complex chemical reasoning tasks is unlikely. 
%

We therefore argue that \textit{molecular representations should first be aligned with structure-grounded language descriptions},
with chemical knowledge and functional reasoning then encoded within the language model through large-scale training.
This resonates with vision--language modeling: image descriptions focus on faithfully describing the \textit{observed visual content}, as illustrated in Fig.~\ref{fig:motivation_illustration}, with the visual module serving primarily as a perception or generation component, while deeper interpretation and reasoning are delegated to the LLM backbone~\citep{survey_mllm}.

Molecule--language alignment thus demands extensive structure-grounded descriptions, as in VLMs~\citep{laion}.
However, curating molecular description datasets presents  different challenges.
First, subtle structural changes can dramatically alter properties and reaction mechanisms~\citep{stereochemical_drug_design,advanced_organic_chemistry}, requiring descriptions to be \textit{complete and unambiguous}.
Second, unlike the image domain, no large-scale molecular structure descriptions readily exist.
Given the complexity of molecular structures, including intricate ring topologies, nested substituents, and stereochemical configurations, human annotation at scale is also impractical.

We address this challenge by developing an automated annotation framework. 
Building upon a rule-based IUPAC name parser, OPSIN~\citep{opsin}, we invest considerable engineering to construct enriched XML metadata encoding molecular structure.
This metadata is injected into LLMs to guide  generation of  descriptions that preserve complete structural details, including
ring topology, substituent positioning, attachment relations, labeling semantics, and stereochemical information.
Using this pipeline, we curate MolLangData, a dataset of $\num{163111}$ molecule--description pairs. 
Rigorous validation on a $\num{2000}$-sample subset by reconstructing the exact structures from descriptions, involving substantial expert effort, 
demonstrates $98.6\%$ annotation precision.
Then, we train a $4$B-parameter LLM on MolLangData via large-scale reinforcement learning  for language-conditional molecule generation.
The training improves the base model's generation accuracy from $12.53\%$ to $84.99\%$. 
Though much smaller, our model significantly outperforms most evaluated frontier LLMs.
Meanwhile, the  model shows strong token efficiency, reducing token usage by $10\times$ compared with open-weight LLMs.
Together, the quality and utility of MolLangData are well justified, establishing a reliable foundation for molecule--language alignment toward advancing chemical reasoning.

\section{Task Formulation and Challenges}
We consider the task of generating a large-scale dataset of molecular structure descriptions.
Specifically, each entry consists of a molecular structure paired with a corresponding structural description, denoted $(\mathcal{M}, \mathcal{T})$.
The description $\mathcal{T}$ is  \textit{complete and unambiguous} such that the molecule $\mathcal{M}$ can be uniquely determined from it.
Conversely,  a given molecular structure may admit multiple valid descriptions due to linguistic and chemical terminology variation,  
but all   should be semantically equivalent.
Generating such a dataset at scale presents several challenges:

	\textbf{Challenge 1: Appropriate Level of Structural Detail}
	%
	At one extreme, a description could enumerate all atoms, bond types, connectivities, and stereochemical relationships.
	Although such descriptions can be generated by traversing the molecular graph, they become verbose even for molecules of moderate size. 
	More importantly, molecular properties and chemical reactions are associated with functional groups and scaffold-level motifs rather than isolated atoms. 
	Purely atom-level descriptions therefore fail to align structure with the abstractions most useful for downstream reasoning.
	At the other extreme, descriptions that only summarize functional groups or scaffold types may omit critical structural details. 
	Subtle differences in ring topologies, attachment positions, or stereochemistry can substantially change physicochemical properties, biological activity, toxicity, or reaction mechanisms~\citep{stereochemical_drug_design,advanced_organic_chemistry}, and cannot be ignored.
	An effective description must therefore balance sufficient structural specificity with chemically meaningful abstractions.
	
	\textbf{Challenge 2: Scalability of Dataset Construction} Molecule--language modeling  requires datasets containing hundreds of thousands to millions of molecule--description pairs.
	According to \citet{mollangbench}, annotating and validating a single structural description can require up to $\qty{1}{h}$ of expert effort.
	%
	Even hybrid pipelines combining LLM-assisted generation with human validation remain impractical at this scale.
	Dataset construction must therefore rely on fully automated generation. 
	
	\textbf{Challenge 3: Insufficient Structural Information} LLMs are increasingly used to scale annotation~\citep{llm4annotation}, but generating accurate  structural descriptions  depends  on complete and precise structural information, which is not conveyed by common molecular representations.
	Linear representations such as SMILES~\citep{smiles}
	specify atom-wise connectivity through grammar rules and traversal order, but do not encode structural abstractions.
	LLMs must instead infer abstractions from atom-level sequences, which we and \citet{mollangbench} find to be error-prone, particularly for complex ring topologies and long side chains.
	In contrast, IUPAC nomenclature more closely reflects how chemists describe molecular structures.
	However, 
	correct interpretation of IUPAC names requires extensive nomenclature knowledge, including tokenization conventions, locant assignment, ring labeling schemes, and fusion, bridged, and spiro descriptors.
	As illustrated in Fig.~\ref{fig:main_example}, interpreting the name requires understanding how to tokenize, where and in what order to parse, what fusion letters mean (e.g., the `e' in fused-ring notation), and how to assign locants after complex ring fusion, bridging, or spiro structures. 
	The IUPAC Blue Book~\citep{iupac_bluebook} takes over $\num{1100}$ pages on organic compound nomenclature. 
	Consequently, generating descriptions directly from IUPAC names remains unreliable, as demonstrated in Appendix Ablation Study~\ref{sec:ablation_study}.

\section{Methodology of Dataset Generation}
Building on OPSIN~\citep{opsin}, a rule-based system for interpreting chemical nomenclature and reconstructing structures,
we develop an automated pipeline that substantially adapts and extends its intermediate representations to produce XML metadata encoding molecular structure.
This metadata is used to guide LLMs in generating complete and unambiguous structure descriptions.
Below, we first briefly introduce OPSIN and analyze the limitations of its native parse tree for description generation, then present our approach for constructing enriched structural metadata, and finally describe the prompt design and filtering strategies used to improve generation accuracy.

\subsection{Limitations of OPSIN for Description Generation}
%
%
Given a chemical name, OPSIN performs grammatical tokenization (e.g., \textit{propan-2-ol} $\rightarrow$ [\textit{prop, an, -, 2-, ol}]) and 
assigns each token a semantic role (e.g., root structures, substituents, locants, suffixes, stereochemistry).
The tokens and their  attributes are organized into an XML parse tree representing intermediate structural interpretations.
The tree is then operated through a staged pipeline that generates, processes, and assembles structural components into a molecular graph, output as CML~\citep{cml}, InChI~\citep{inchi}, or SMILES~\citep{smiles}. 

However, OPSIN's XML is designed as an internal representation rather than a complete, accurate structural encoding, making it unsuitable for guiding description generation.
%
%
%
Specifically, (i) structural elements such as substituents, locants, suffixes, and stereochemical descriptors are often not positioned to reflect their true attachment or affiliation in the final structure;
(ii) many intermediate elements are discarded once used, leaving the final XML representation incomplete;
(iii) topological relationships, such as fused-ring connectivity, bridged and spiro junctions, implicit locants, and atom labeling schemes, are handled internally and never recorded; and
(iv) some  components (e.g., von Baeyer spiro systems) are resolved directly into SMILES, from which alone LLMs struggle to interpret structure.
%
Enriched structural metadata tailored for description generation is thus  needed.

\subsection{Enriched Structural Metadata Construction}

\begin{figure}[!t]
	\includegraphics[width=\linewidth]{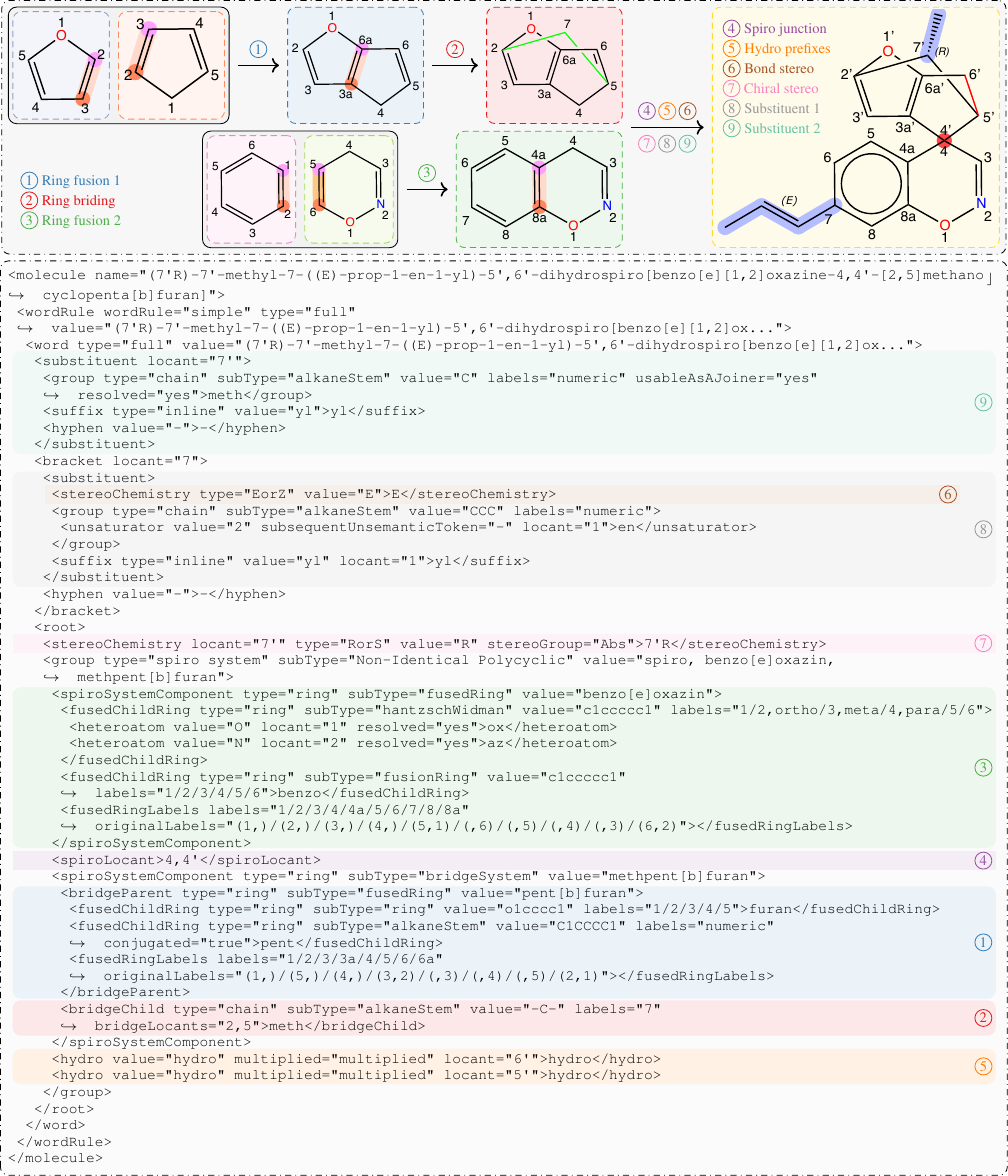}
	\caption{
		Decomposition of an example molecule from basic components to the complete structure, along with the XML metadata constructed by our approach.
		The native OPSIN XML output and generated natural-language description are provided in Appendix Figs.~\ref{fig:running_example_opsin_xml} and~\ref{fig:running_example_descriptions}, respectively.
	}
	
	\vspace{-0.25in}
	\label{fig:main_example}
\end{figure}

We fully leverage OPSIN's grammatical tokenization and initial XML parse tree construction.
We perform substantial engineering to transform the parse tree into a metadata representation suitable for guiding molecular structure description generation.
Importantly, these modifications do not alter OPSIN's native structure construction process.
Our tool serves as an adaptive extension of OPSIN, preserving full compatibility with chemical name-to-structure conversion while providing a semantically complete and well-organized structural representation.
This design facilitates compatibility with future OPSIN updates and a principled sanity check by comparing converted structures with those from the native OPSIN pipeline, ensuring the reliability of generated metadata.

A systematic redesign of OPSIN is impractical, as both IUPAC nomenclature and OPSIN are highly complex; despite being actively developed and maintained for over a decade, OPSIN continues to encounter unresolved special cases.
We therefore adopt a trial-and-error engineering strategy.
We iteratively test and refine our metadata construction on
over $\num{2000}$ molecules covering diverse and challenging cases, including but not limited to, complex fused, bridged, and spiro ring systems; Hantzsch-Widman heterocycles; multiple stereochemical configurations; amino acid derivatives; and a wide range of prefixes, infixes, and suffixes.
%
%
Using the representative example in Fig.~\ref{fig:main_example} and the corresponding native OPSIN XML  in Appendix Fig.~\ref{fig:running_example_opsin_xml},
we summarize the key modifications:

\textit{Augment topology and labeling information.}
Molecules containing complex ring systems
comprise over $30\%$ of entries in public repositories such as PubChem~\citep{pubchem}.
In native OPSIN, ring-construction information  
and the resulting atom labeling schemes are processed internally but not written in the parse tree.
This missing information is critical both for accurately describing ring topology and correctly determining the attachment locants for subsequent structural components.
We therefore augment the XML  to encode the missing topology and atom labeling information.

For fused rings, we first record each base ring as a \texttt{fusedChildRing} element, storing its atom set (\texttt{value} attribute) and original atom labels (\texttt{labels} attribute).
A \texttt{fusedRingLabels} structure then specifies
both the fusion junctions between the base rings (\texttt{originalLabels} attribute) and the new atom labeling scheme of the fused system (\texttt{labels} attribute).
In the \textit{benzo[e][1,2]oxazine} fusion (process \textcircled{\scriptsize 3}), 
\textit{oxazine} and the \textit{benzo}  rings are fused by sharing two atom pairs, $(5,1)$ and $(6,2)$:
\textit{oxazine} atoms 5 and 6 are fused with \textit{benzo} atoms 1 and 2, respectively.
%
After fusion, these junction atoms are relabeled as $4a$ and $8a$. 
The complete fusion process is captured in the augmented metadata.
Bridged and spiro ring systems are handled analogously.
These representations support cascaded multi-ring fusions and nested topologies, as in the illustrative example.
Semantic definitions for fused, bridged, and spiro systems are provided in Prompts~\ref{prompt:fused_ring_semantics}, \ref{prompt:bridged_ring_semantics}, and \ref{prompt:spiro_ring_semantics} in the Appendix~\ref{appdix:generation_prompt}.
	
\textit{Retain critical elements.}
We modify OPSIN to retain structurally meaningful elements that are otherwise discarded during component generation and structure assembly.
As shown in Fig.~\ref{fig:main_example}, stereochemical information (\textcircled{\scriptsize 6} and \textcircled{\scriptsize 7}), unsaturation descriptors such as the \textit{en} modifier of the \textit{prop} substituent (\textcircled{\scriptsize 8}), hydro prefixes (\textcircled{\scriptsize 5}), and core ring descriptors, including heteroatom information in the \textit{oxazine} ring (\textcircled{\scriptsize 3}), are removed in the native OPSIN parse tree but explicitly retained in our metadata.
In addition, 
fused, bridged, and spiro ring information, which native OPSIN never creates and reduces to a generic systematic scaffold name,
is fully augmented in our XML representation (\textcircled{\scriptsize 1}–\textcircled{\scriptsize 4}), as described above.
All retained elements are further marked and post-processed to prevent re-interpretation or unintended modification in subsequent OPSIN processing stages.

\textit{Rearrange elements to reflect affiliation and connectivity.}
The native OPSIN XML organizes elements according to IUPAC token order, whereas actual modification and attachment relationships are resolved through rule-based inference during structure assembly.
Thus, locants, prefixes, and stereochemical descriptors may appear detached from the structural components they actually modify.
For example, in \textit{(E)-5-(prop-1-en-1-yl)non-3-ene}, the `\textit{E}' descriptor applies to the \textit{non-3-ene} backbone, whereas in \textit{(E)-5-(prop-1-en-1-yl)non-1-ene}, it instead applies to the \textit{prop-1-en-1-yl} substituent.
A similar issue arises in the illustrative example, where prefixes such as hydrogenation modifiers should apply to the fully resolved fused or spiro scaffold rather than to early name fragments.
However, the native OPSIN parse tree does not encode these relationships.
We systematically relocate prefixes, locants, stereochemical descriptors, and substituents in the XML tree so that each element is attached to the structural component it modifies in the final molecular graph.
	
\textit{Miscellaneous corner cases.} In addition to the major structural deficiencies discussed above, our practical implementation required substantial engineering effort to address many additional corner cases where native OPSIN metadata is incomplete or difficult for LLMs to interpret.
For example, explicit heteroatom positions in Hantzsch-Widman systems (e.g., \textit{1,2-oxazole}) are not encoded in the native XML representation.
Similarly, for von Baeyer spiro systems, OPSIN directly resolves name fragments into SMILES strings via rule-based functions, 
which are effective for structure construction but provide limited and unreliable guidance for LLM-based description generation.
Although corner cases remain, our iterative refinement enables construction of complete and structured metadata for the vast majority of molecules encountered in practice, as supported by our validation results showing that no description errors result from incomplete or incorrect metadata.

\subsection{Prompt Design and Atom-Matching Filtering}
In addition to our constructed metadata, the LLM is provided with the corresponding SMILES  and IUPAC name,
and is instructed to generate a self-contained, natural-language structure description.
The prompt used for description generation is given in Appendix Prompt~\ref{prompt:desp_gen}.
The generated description is required to be sufficiently complete and precise that someone with basic organic chemistry knowledge can reconstruct the structure exactly and unambiguously from the description alone.
The prompt includes  guidance on describing molecular backbones, connectivity, substituents, functional groups, and stereochemistry.
When fused, spiro, or bridged ring systems are present, additional semantic specifications are  injected to guide the interpretation of complex ring topologies.
%

Despite the detailed structural information provided, preliminary experiments show that the LLM occasionally omits atoms in long side chains or chain linkers.
To mitigate this, we incorporate an automated atom-matching filtering step during description generation.
In addition to description generation, the LLM is prompted to count  the total non-hydrogen atoms based solely  on its generated description. 
Descriptions whose reported atom counts do not match the ground truth are  discarded.

More rigorous self-checking mechanisms are possible, such as prompting the LLM to reconstruct the SMILES from the generated description and performing full structure matching.
However, this strategy is not suited for large-scale dataset construction.
First, including SMILES reconstruction within the same prompt introduces a non-trivial task that can interfere with the primary objective of structural description generation;
alternatively, performing reconstruction in a separate LLM call substantially increases cost, making it impractical at scale.
Second, single-pass reconstruction checking suffers from high false rejection rates: even strong models  can reject a large fraction of otherwise correct descriptions, resulting in low annotation efficiency despite near-zero description errors.
We therefore adopt this lightweight yet effective atom-matching filter, which improves description accuracy while preserving scalability,  as demonstrated in the ablation study (Sec.~\ref{sec:ablation_study} in the Appendix).



\section{Dataset Collection and Validation}

\subsection{Dataset Collection}
Chemical space is extremely large~\citep{chemical_space}; 
%
even public molecule repositories contain hundreds of millions to billions of molecules~\citep{pubchem, zinc20}. 
We randomly sample $\num{200000}$ molecules from PubChem~\citep{pubchem} as candidates for dataset construction.
%
We then apply initial data-quality and consistency checks before description generation.
%
After filtering, {$\num{167416}$} molecules remain and are passed to the LLM for description generation. 
Multi-component entries account for $88.7\%$ of excluded candidates. 
OPSIN and our tool natively support these entries, and each component can also be processed separately.
We exclude them only to simplify validation.
%
The remaining exclusions are mainly due to missing IUPAC names or mismatches caused by errors in the PubChem SMILES records. 
Fewer than $1\%$ of candidates are excluded due to OPSIN parsing limitations.
Detailed filtering statistics are provided in Appendix~\ref{appendix:filter_statistics}.




While structural metadata provides sufficient information, 
accurate generation still requires the LLM to faithfully translate it into precise textual descriptions.
We therefore use GPT-5.2~\citep{openai5_2} as the generation model (the most capable model available at the time of dataset curation).
We group molecules into three generation difficulty levels, and use high reasoning effort for easy molecules and xhigh for medium and hard molecules.
The difficulty criteria  and the model snapshots used for generation are provided in  Appendix~\ref{appendix:difficulty_category} and \ref{appdix:model_snapshots}, respectively.
Note that \emph{the difficulty level does not necessarily reflect molecular complexity}; molecules in the easy category may still be structurally intricate, as illustrated in Appendix Fig.~\ref{fig:easy complex}.
%
%
As reported in Appendix~\ref{appendix:data_generation_cost}, curating the dataset with GPT-5.2 incurred over \text{\$}$20$k  in API costs. 
While higher-capacity models like GPT-5.5 can produce more accurate descriptions, they cost $2\times$ as much. 
The choice of generation model and reasoning effort for each difficulty level balances generation quality and  practical budget constraints.

We additionally prompt the LLM to count non-hydrogen atoms to filter potentially erroneous descriptions.
After filtering, the final dataset consists of {$\num{163111}$} molecule--description pairs, 
with the distribution across difficulty levels summarized in~Table~\ref{tab:data_generation_stat}.
%
%
On average, each molecule contains $26.24$ heavy atoms, and its corresponding description consists of  $\num{261}$ words. 
Representative examples and additional dataset statistics are provided in  Appendix~\ref{sec:data_samples} and ~\ref{sec:appendix_dataset_statistics}, respectively.

\subsection{Dataset Validation}
\label{sec:data_validation}
To assess reliability,
we validate a random subset of $\num{2000}$ from the dataset that pass the atom-matching check,  
comprising $\num{1317}$, $\num{496}$, and $\num{187}$ easy, medium, and hard samples, respectively.

	\textbf{Validation Process} Human validation of molecular descriptions is time-consuming;
	based on our records, validating a single sample requires approximately $\qty{11.7}{\minute}$ on average, slightly above the $\qty{10}{\minute}$ reported by \citet{mollangbench}.
	Fully validating even the $\num{2000}$-sample subset thus requires about $\num{390}$ human-hours for a single round, excluding additional validation for ablation studies. 

	To evaluate more samples while keeping human validation practical, 
	we combine LLM-based and human validation.
	For each sample, we prompt GPT-5.2, GLM-5.1, and DeepSeek-V4-Pro to reconstruct the molecule in SMILES solely from its description (Appendix Prompt~\ref{prompt:molecule_generation}), with up to three independent attempts per model.
	A sample is accepted only when \emph{all} three LLMs exactly reconstruct the ground-truth structure.
	We expect near-zero false positives because exact reconstruction from an incorrect or ambiguous description by chance is unlikely.
	Even for the $\num{626}$ samples accepted by only one or two LLMs and thus rejected by automatic validation, later human validation confirms unambiguous descriptions and exact reconstruction.
	Requiring agreement among all three LLMs provides an additional safeguard.
	Human validation of $200$ randomly selected samples accepted by all three also finds no false positives.
	Appendix~\ref{sec:false_positive_validation} provides the full empirical analysis.
	The LLMs successfully validate $62.8\%$ of samples, 
	reducing the number requiring human validation to $\num{744}$.
	
	However, LLM validation may produce false negatives.
	All samples failing LLM validation undergo human validation, where each sample is independently evaluated by up to two expert validators.
	Validators are provided only with the textual description and reconstruct the molecule using chemical drawing software;
	their reconstructed structures are submitted to an internal validation application that automatically checks them against the ground truth.
	Each validator is allowed up to three attempts.
	A sample is considered valid if either expert confirms that the description is \emph{unambiguous} and that they successfully reconstruct the \emph{exact}  structure.
	A sample is labeled incorrect only if both validators fail to reconstruct it, mitigating the risk of individual oversight.
    Although LLM validation reduces the human workload, expert validation still requires approximately $\qty{160}{\hour}$. 
    We additionally spend around $\qty{110}{\hour}$ on validation for the false-positive study and the ablation studies described later.
	
\begin{wraptable}{r}{0.53\linewidth}
	\scriptsize
	\caption{Dataset statistics across generation difficulty levels, including collected samples, validation subset size, and validation precision ($\%$).
		For collected and validation samples, each entry is reported as the sample count with its proportion ($\%$) relative to the total dataset or total validation set, respectively.}
\label{tab:data_generation_stat}
\definecolor{easybg}{RGB}{245,245,245}
\definecolor{mediumbg}{RGB}{235,235,235}
\definecolor{hardbg}{RGB}{225,225,225}
\definecolor{highbg}{RGB}{245,245,245}
\definecolor{xhighbg}{RGB}{235,235,235}
\definecolor{mainbg}{RGB}{235,240,245}
\definecolor{tableheadcolor}{gray}{0.92}
	\begin{tabular}{@{}p{0.09\linewidth}  >{}p{0.26\linewidth}  >{}p{0.24\linewidth}  >{}p{0.24\linewidth}@{}}
		\toprule
		\rowcolor{tableheadcolor}
		Difficulty  & 
		{Collected samples} &
		{Validated samples} &
		{Correct samples}  \\
		\midrule
		Easy &
		$\num{106402}$ ($65.2\%$)&
		$\num{1317}$ ($65.8\%$) &
		$\num{1300}$ ($98.7\%$) \\
		\midrule
		Medium &
		$\num{41412}$ ($25.4\%$)&
		$\num{496}$ ($24.8\%$) &
		$\num{492}$ ($99.2\%$) \\
		\midrule
		Hard &
		$\num{15297}$ ($9.4\%$)&
		$\num{187}$ ($9.4\%$) &
		$\num{180}$ ($96.3\%$) \\
		\midrule
		\textbf{Overall} &
		$\mathbf{163,111}$ &
		$\mathbf{2,000}$ &
		$\mathbf{1,972}$ ($98.6\%$) \\
		\bottomrule
	\end{tabular}%
\end{wraptable}

	\textbf{Validation Results}
	We report description precision on the validation set across difficulty levels in Table~\ref{tab:data_generation_stat}, with additional validation statistics in Appendix~\ref{sec:validation_stat}.
	The overall precision of validated samples is {$98.6\%$}.
	
	We  examine all incorrect descriptions and confirm that structure parsing and metadata construction are correct;
	errors therefore arise solely from LLM description generation.
	In all incorrect cases, the errors are minor:  descriptions contain all scaffolds and substituents, with most structural elements correctly connected and only localized issues, such as misplaced substituents or imprecise ring fusion descriptions.
	%
	%
	Overall, the low error rate and minor errors
	demonstrate the reliability of the generation pipeline and the resulting dataset.

\subsection{Ablation Study}
We perform ablation studies to analyze the impact of our constructed molecular-structure XML metadata, atom-matching filtering, and the choice of generation models.
Each component meaningfully affects description precision.
Full results and analysis are provided in Appendix~\ref{sec:ablation_study}.


\section{Language-Conditional Molecule Generation}
\label{sec:language_conditioned_generation}

Molecule--language modeling is bidirectional between molecular structures and natural-language descriptions,
including both perception and generation.
This section showcases large-scale training on our curated dataset for language-conditional molecule generation:
given a detailed and unambiguous structural description, the model is to generate the exact corresponding molecular structure.

One direct application of this model is integration with chemical drawing software, allowing users to create structures from language or voice instructions instead of drawing atoms and bonds manually.
%
More importantly, such a model could support chemical design, optimization, and synthesis planning in LLM-centric AI systems by translating design intent into concrete molecular structures.


\newcommand{\MolGenTrajectoryWidth}{0.6\textwidth}
\newcommand{\MolGenTrajectoryHeight}{3.8cm}
\newcommand{\MolGenSummaryWidth}{0.28\textwidth}
\newcommand{\MolGenSummaryHeight}{\MolGenTrajectoryHeight}
\newcommand{\MolGenSummaryGap}{2pt}
\newif\ifMolGenShowModelNames
\MolGenShowModelNamestrue
\newcommand{\MolGenModelLabel}[3]{%
    \ifMolGenShowModelNames
        \node[font=\tiny\rmfamily, anchor=#1] at (axis cs:#2) {#3};%
    \fi
}

\begin{figure*}[t]
\centering
\begin{tikzpicture}
\begin{axis}[
    name=trainingplot,
    width=\MolGenTrajectoryWidth,
    height=\MolGenTrajectoryHeight,
    scale only axis,
    xmin=0, xmax=100,
    axis x line*=bottom,
    xtick={0,6.667,13.333,20,46.667,73.333,100},
    xticklabels=\empty,
    xtick style={draw=black!75},
    clip=false,
    ylabel={Exact-match rate ($\%$)},
    ylabel style={font=\scriptsize, blue!65!black, at={(axis description cs:0.10,0.5)}, anchor=south},
    ymin=10, ymax=100,
    ytick={10,20,40,60,80,100},
    ytick pos=left,
    tick label style={font=\tiny, blue!65!black},
    grid=major,
    grid style={gray!20},
    axis line style={black!65},
    y axis line style={blue!65!black, -},
]

\begin{pgfonlayer}{background}
\path[fill=cyan!2] (axis cs:0,10) rectangle (axis cs:20,100);
\path[fill=green!2] (axis cs:20,10) rectangle (axis cs:100,100);
\end{pgfonlayer}
\addplot[black!60, dashed, forget plot] coordinates {(20,10) (20,100)};

\addplot[blue!40, dashed, forget plot] coordinates {(0,46.42) (100,46.42)};
\addplot[blue!40, dashed, forget plot] coordinates {(0,49.39) (100,49.39)};
\addplot[blue!40, dashed, forget plot] coordinates {(0,63.54) (100,63.54)};
\addplot[blue!40, dashed, forget plot] coordinates {(0,65.18) (100,65.18)};
\addplot[blue!40, dashed, forget plot] coordinates {(0,68.21) (100,68.21)};
\addplot[blue!40, dashed, forget plot] coordinates {(0,81.95) (100,81.95)};
\addplot[blue!40, dashed, forget plot] coordinates {(0,81.44) (100,81.44)};
\addplot[blue!40, dashed, forget plot] coordinates {(0,92.95) (100,92.95)};
\addplot[blue!40, dashed, forget plot] coordinates {(0,94.27) (100,94.27)};
\begin{pgfonlayer}{foreground}
\node[font=\tiny\rmfamily, text=blue!65!black, inner sep=0.6pt, anchor=west]
    at (axis cs:26,43.52) {qwen3.6-35b};
\node[font=\tiny\rmfamily, text=blue!65!black, inner sep=0.6pt, anchor=west]
    at (axis cs:36,51.59) {gpt-5};
\node[font=\tiny\rmfamily, text=blue!65!black, inner sep=0.6pt, anchor=west]
    at (axis cs:40,60.54) {qwen3.8-max};
\node[font=\tiny\rmfamily, text=blue!65!black, inner sep=0.6pt, anchor=west]
    at (axis cs:48,65.18) {glm-5.2};
\node[font=\tiny\rmfamily, text=blue!65!black, inner sep=0.6pt, anchor=west]
    at (axis cs:54,70.21) {deepseek-v4-pro};
\node[font=\tiny\rmfamily, text=blue!65!black, inner sep=0.6pt, anchor=west] (gpt54label)
    at (axis cs:66,87.05) {gpt-5.4};
\draw[black, line width=0.5pt,
    -{Stealth[length=1.5pt,width=1pt]}]
    (axis cs:72,86.05) -- (axis cs:77.3,82.2);
\node[font=\tiny\rmfamily, text=blue!65!black, inner sep=0.6pt, anchor=west]
    at (axis cs:60,78.44) {gemini-3.7};
\node[font=\tiny\rmfamily, text=blue!65!black, inner sep=0.6pt, anchor=west]
    at (axis cs:76,90.35) {gpt-5.5};
\node[font=\tiny\rmfamily, text=blue!65!black, inner sep=0.6pt, anchor=west]
    at (axis cs:82,96.27) {gpt-5.6-sol};
\end{pgfonlayer}

\addplot[semithick, blue!70!black, mark=*, mark size=0.55pt] coordinates {
    (0,12.53) (6.667,37.32) (13.333,50.25) (20,59.84)
};
\addplot[semithick, blue!70!black, forget plot]
    coordinates {(20,59.84) (20.267,58.42)};
\addplot[semithick, blue!70!black] table[
    col sep=comma,
    x expr={20 + 80*\thisrow{rollout_step}/1500},
    y expr={100*\thisrow{exact_at_1}},
] {figures/training_trajectory_rl_mollangdata.csv};

\node[font=\tiny, anchor=north] at ([yshift=1pt]axis cs:0,10) {0};
\node[font=\tiny, anchor=north] at ([yshift=1pt]axis cs:6.667,10) {1};
\node[font=\tiny, anchor=north] at ([yshift=1pt]axis cs:13.333,10) {2};
\node[font=\tiny, anchor=north] at ([yshift=1pt]axis cs:20,10) {3 (0)};
\node[font=\scriptsize, anchor=north]
    at ([yshift=-7pt]axis cs:10,10) {SFT epoch};
\node[font=\tiny, anchor=north] at ([yshift=1pt]axis cs:46.667,10) {500};
\node[font=\tiny, anchor=north] at ([yshift=1pt]axis cs:73.333,10) {1k};
\node[font=\tiny, anchor=north] at ([yshift=1pt]axis cs:98,10) {1.5k};
\node[font=\scriptsize, anchor=north]
    at ([yshift=-7pt]axis cs:60,10) {RL rollout step (1 rollout step = 16$\times$ optimization steps)};

\end{axis}

\begin{axis}[
    at={(trainingplot.south west)},
    anchor=south west,
    width=\MolGenTrajectoryWidth,
    height=\MolGenTrajectoryHeight,
    scale only axis,
    xmin=0, xmax=100,
    xtick={0,6.667,13.333,20,46.667,73.333,100},
    ymin=700, ymax=60000,
    ymode=log,
    ytick={1000,2000,5000,10000,20000,50000},
    yticklabels={},
    axis x line=none,
    axis x line*=bottom,
    axis y line*=right,
    yticklabel style={font=\tiny, red!70!black, text width=2.2em,
        align=center, anchor=west, xshift=-7pt},
    axis line style={red!70!black, -},
]
\addplot[red!70!black, densely dotted, forget plot] coordinates {(0,51451) (100,51451)};
\addplot[red!70!black, densely dotted, forget plot] coordinates {(0,7539) (100,7539)};
\addplot[red!70!black, densely dotted, forget plot] coordinates {(0,6707) (100,6707)};
\addplot[red!70!black, densely dotted, forget plot] coordinates {(0,10912) (100,10912)};
\addplot[red!70!black, densely dotted, forget plot] coordinates {(0,18469) (100,18469)};
\addplot[red!70!black, densely dotted, forget plot] coordinates {(0,3700) (100,3700)};
\addplot[red!70!black, densely dotted, forget plot] coordinates {(0,3495) (100,3495)};
\addplot[red!70!black, densely dotted, forget plot] coordinates {(0,2527) (100,2527)};
\addplot[red!70!black, densely dotted, forget plot] coordinates {(0,2623) (100,2623)};
\addplot[red!70!black, densely dotted, forget plot] coordinates {(0,1219) (100,1219)};
\addplot[red!70!black, densely dotted, forget plot] coordinates {(0,1058) (100,1058)};

\addplot[semithick, red!75!black, mark=square*, mark size=0.55pt] coordinates {
    (0,19338) (6.667,14103) (13.333,14479) (20,14304)
};
\addplot[semithick, red!75!black, forget plot]
    coordinates {(20,14304) (20.267,15412)};
\addplot[semithick, red!75!black] table[
    col sep=comma,
    x expr={20 + 80*\thisrow{rollout_step}/1500},
    y=avg_output_tokens,
] {figures/training_trajectory_rl_mollangdata.csv};
\end{axis}

\begin{axis}[
    at={(trainingplot.south west)},
    anchor=south west,
    width=\MolGenTrajectoryWidth,
    height=\MolGenTrajectoryHeight,
    scale only axis,
    xmin=0, xmax=100,
    ymin=0, ymax=1,
    xtick={0,6.667,13.333,20,46.667,73.333,100},
    xticklabels=\empty,
    axis x line*=top,
    axis y line=none,
    clip=false,
    ytick=\empty,
    tick label style={font=\scriptsize},
    xlabel style={font=\scriptsize, at={(axis description cs:0.5,1.23)}, anchor=south},
]

\end{axis}

\begin{axis}[
    name=summaryplot,
    at={(trainingplot.north east)},
    anchor=north west,
    xshift=\MolGenSummaryGap,
    width=\MolGenSummaryWidth,
    height=\MolGenSummaryHeight,
    scale only axis,
    xmin=40, xmax=100,
    xtick={40,50,60,70,80,90,100},
    xticklabels={,50,60,70,80,90,100},
    ymode=log,
    ymin=700, ymax=60000,
    ytick={1000,2000,5000,10000,20000,50000},
    yticklabels={1k,2k,5k,10k,20k,50k},
    yticklabel style={font=\tiny, red!70!black, 
     anchor=west, xshift=-2pt},
    yticklabel pos=right,
    clip=false,
    xlabel={Exact-match rate ($\%$)},
    xlabel style={font=\scriptsize, blue!65!black, yshift=8pt},
    x tick label style={font=\tiny, blue!65!black, yshift=1pt},
    grid=major,
    grid style={gray!20},
    axis line style={black!65, -},
	separate axis lines=true,
	axis line style={-},
	x axis line style={blue!65!black},
	y axis line style={red!70!black},
    axis background/.style={
        path picture={
            \node[anchor=south west, inner sep=0pt, opacity=0.35] at
                (path picture bounding box.south west) {\includegraphics[
                    width=\pgfkeysvalueof{/pgfplots/width},
                    height=\pgfkeysvalueof{/pgfplots/height}
                ]{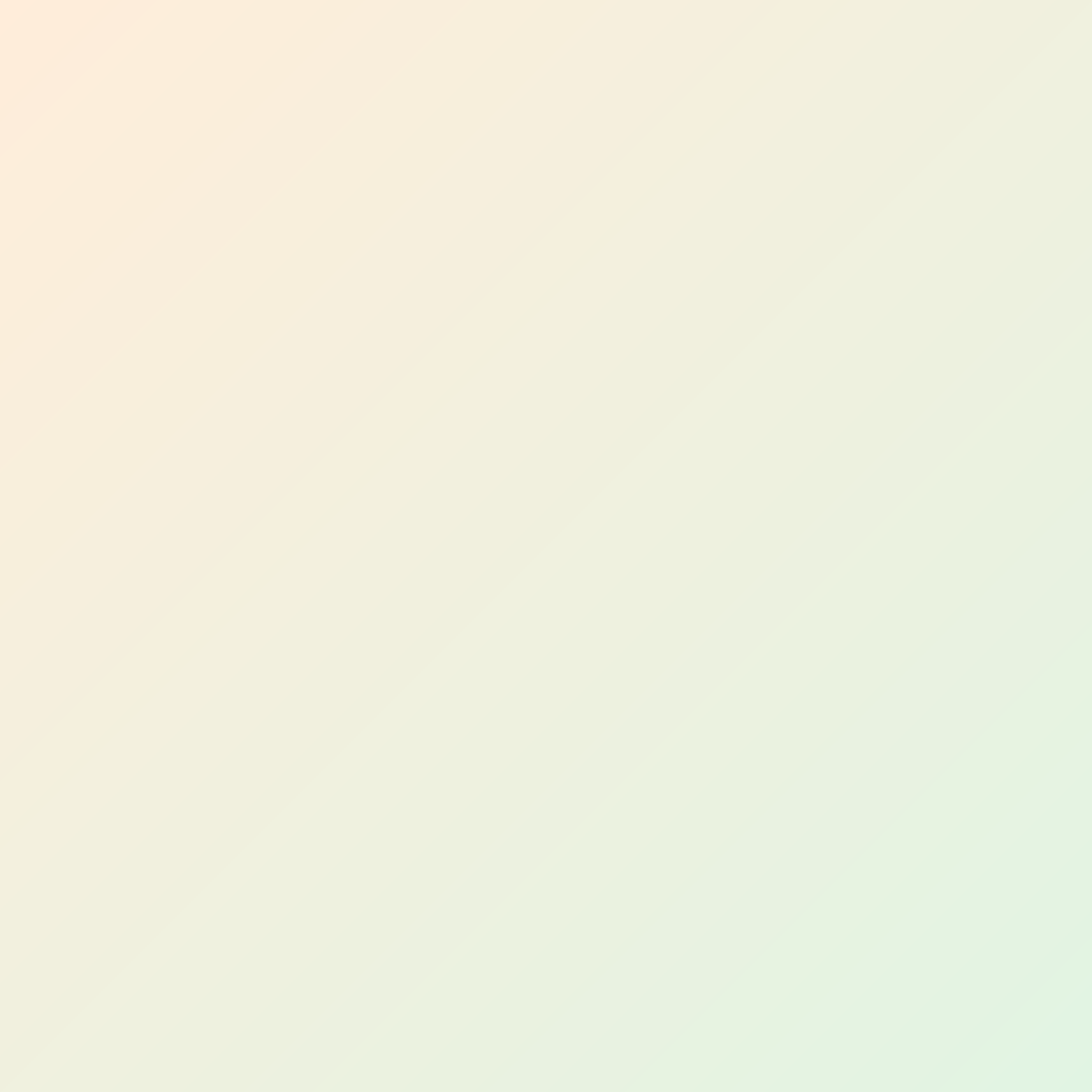}};
        },
    },
]
\coordinate (summaryQwen36) at (axis cs:46.42,51451);
\coordinate (summaryGpt5) at (axis cs:49.39,7539);
\coordinate (summaryQwen38Max) at (axis cs:63.54,6707);
\coordinate (summaryGlm52) at (axis cs:65.18,10912);
\coordinate (summaryDeepSeek) at (axis cs:68.21,18469);
\coordinate (summaryGpt52) at (axis cs:68.56,3700);
\coordinate (summaryGpt54) at (axis cs:81.95,3495);
\coordinate (summaryGemini) at (axis cs:81.44,2527);
\coordinate (summaryGpt55) at (axis cs:92.95,2623);
\coordinate (summaryTerra) at (axis cs:73.48,1219);
\coordinate (summarySol) at (axis cs:94.27,1058);
\node[inner sep=0pt] at (axis cs:46.42,51451)
    {\includegraphics[width=5.2pt]{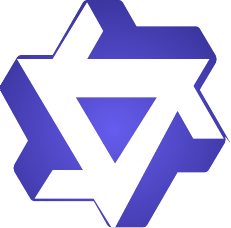}};
\MolGenModelLabel{west}{46.42,51451}{qwen3.6-35b}
\node[inner sep=0pt] at (axis cs:49.39,7539)
    {\includegraphics[width=5.2pt]{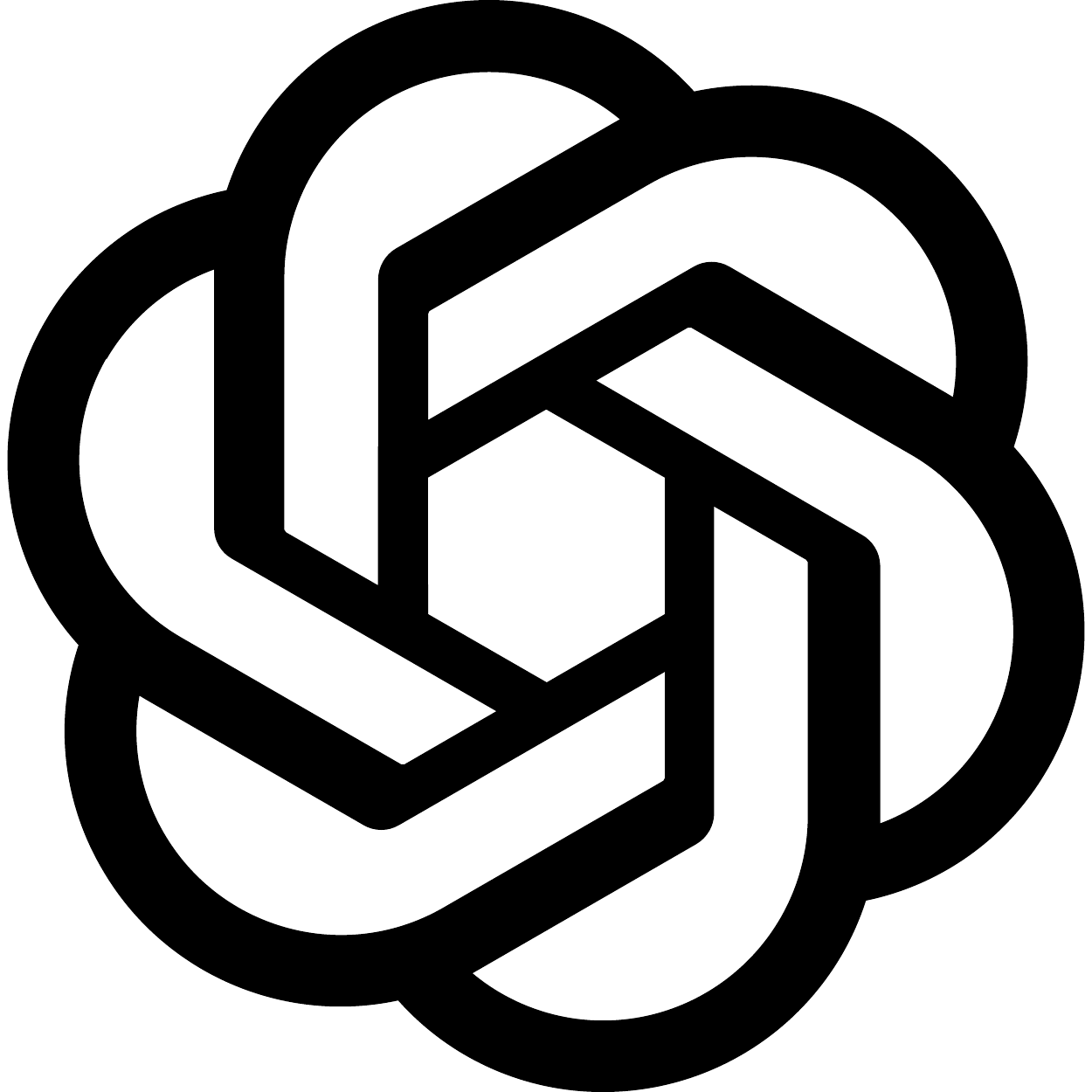}};
\MolGenModelLabel{west}{49.39,7539}{gpt-5}
\node[inner sep=0pt] at (axis cs:63.54,6707)
    {\includegraphics[width=5.2pt]{figures/qwen.pdf}};
\MolGenModelLabel{west}{63.54,6707}{qwen3.8-max}
\node[inner sep=0pt] at (axis cs:65.18,10912)
    {\includegraphics[width=5.2pt]{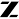}};
\MolGenModelLabel{west}{65.18,10912}{glm-5.2}
\node[inner sep=0pt] at (axis cs:68.21,18469)
    {\includegraphics[width=5.2pt]{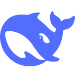}};
\MolGenModelLabel{south}{68.21,18469}{deepseek-v4-pro}
\node[inner sep=0pt] at (axis cs:68.56,3700)
    {\includegraphics[width=5.2pt]{figures/openai.pdf}};
\MolGenModelLabel{east}{68.56,4200}{gpt-5.2}
\node[inner sep=0pt] at (axis cs:81.95,3495)
    {\includegraphics[width=5.2pt]{figures/openai.pdf}};
\MolGenModelLabel{south}{81.95,3495}{gpt-5.4}
\node[inner sep=0pt] at (axis cs:81.44,2527)
    {\includegraphics[width=5.2pt]{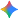}};
\MolGenModelLabel{north}{81.44,2527}{gemini-3.7}
\node[inner sep=0pt] at (axis cs:92.95,2623)
    {\includegraphics[width=5.2pt]{figures/openai.pdf}};
\MolGenModelLabel{south}{92.95,2623}{gpt-5.5}
\node[inner sep=0pt] at (axis cs:73.48,1219)
    {\includegraphics[width=5.2pt]{figures/openai.pdf}};
\MolGenModelLabel{north east}{76.48,1219}{gpt-5.6-terra}
\node[inner sep=0pt] at (axis cs:94.27,1058)
    {\includegraphics[width=5.2pt]{figures/openai.pdf}};
\MolGenModelLabel{north east}{99.27,1058}{gpt-5.6-sol}
\node[inner sep=0pt, font=\small, text=violet!75!black]
    at (axis cs:59.84,14304) {$\bigstar$};
\MolGenModelLabel{east}{58.84,14304}{post-SFT}
\node[inner sep=0pt, font=\small, text=green!65!black]
    at (axis cs:84.99,1264) {$\bigstar$};
\MolGenModelLabel{south west}{83.99,1204}{post-RL}
\node[font=\tiny, text=blue!65!black, anchor=north, xshift=2pt, yshift=1pt]
    at (axis cs:40,700) {40};
\end{axis}
\foreach \endpoint in {summaryQwen36,summaryGpt5,summaryQwen38Max,
    summaryGlm52,summaryDeepSeek,summaryGpt52,summaryGpt54,
    summaryGemini,summaryGpt55,summaryTerra,summarySol} {
    \draw[red!70!black, densely dotted, shorten >=3pt]
        (trainingplot.east |- \endpoint) -- (\endpoint);
}
\node[rotate=-90, anchor=north, font=\scriptsize, text=red!70!black]
at ([xshift=22pt]summaryplot.east) {Avg. output tokens (log scale)};
\end{tikzpicture}
\caption{The left panel shows the training trajectory of exact-match rate and average output tokens on our test set during SFT and RL, with horizontal reference lines for selected LLM baselines. The right panel compares the final post-RL model with these baselines. Full results are in Table~\ref{tab:model_benchmark_results}.}
\label{fig:molecule_generation_training_trajectory}
\vspace{-0.25in}
\end{figure*}

\subsection{Experimental Setup}
Representing molecular structures as SMILES allows a language-only LLM to handle this multimodal task.
We train a LoRA adapter on Qwen3.5-4B~\citep{qwen3.5} using reinforcement learning (RL) to guide the model toward efficient, accurate reasoning for molecule generation.
Since RL training for a small LLM can be difficult to stabilize when the initial policy is weak, we first perform supervised fine-tuning (SFT) before RL, following common practice for RL training of small models~\citep{deepseek}.
We use the same prompt as in the LLM-based reconstruction validation.
Our curated dataset contains $\num{163111}$ samples. 
Of these, $\num{2000}$ samples were validated (Sec.~\ref{sec:data_validation}) and are reserved for later evaluation, while the remaining $\num{161111}$ samples form the training set.

\textbf{Supervised fine-tuning}
Using the training set, we collect outputs from GLM-5.1 and Qwen3.6-35B for the same language-conditional molecule generation task.
Among the generated outputs, $\num{80850}$ exactly match the target molecules.
After removing samples with reasoning traces longer than $\num{40960}$ tokens, we retain $\num{75666}$ samples. 
We use both their reasoning traces and final answers as the SFT training corpus.
The adapter is fine-tuned for $3$ epochs, taking approximately $\qty{19.1}{\hour}$ on  $8$ H100 GPUs.
Appendix~\ref{appdix:sft} provides further details on the SFT dataset and training configurations.

\textbf{Reinforcement learning}
After SFT,  we apply large-scale RL using the decoupled clip and dynamic sampling policy optimization (DAPO)~\citep{dapo}.
The reward function rewards correctly formatted outputs.
It also provides tiered rewards based on whether the SMILES is valid, similar to, or exactly matches the ground-truth, and applies a soft penalty to overlong reasoning traces.
We conduct RL on the full $\num{161111}$-sample training set for $\num{1500}$ rollout steps.
At each rollout step, we sample $512$ prompts and generate $8$ responses per prompt for group-based reward, followed by $16$ optimization steps.
The full RL training takes about $\qty{322.6}{\hour}$ on 
two DGX H100 nodes with $16$ GPUs in total.
More details on the reward formula and RL settings are provided in Appendix~\ref{appdix:rl}.



\subsection{Experimental Results}
Of the $\num{2000}$ validated samples, $\num{1972}$ pass our validation and form the MolLangData test set.
We also evaluate on the expert-written and validated MolLangBench core and extended sets~\citep{mollangbench}, each containing 200 in-the-wild descriptions of molecules with no  overlap with our training set.
%
We evaluate Qwen3.5-4B at different training stages and compare it with 16 much larger LLMs
from the GPT~\citep{openai5_6}, Gemini~\citep{gemini3_7}, DeepSeek~\citep{deepseekv4}, Qwen~\citep{qwen3.5}, and GLM~\citep{glm}  series
\footnote{Claude was evaluated, but safety classifiers refused some prompts;  partial results are in Appendix~\ref{sec:model_benchmark}.}.
%
Fig.~\ref{fig:molecule_generation_training_trajectory} shows our training trajectory on the MolLangData test set, together with representative LLMs, 
in terms of exact-match rate to the target molecule and average output tokens, including reasoning tokens. 
Appendix Table~\ref{tab:model_benchmark_results} reports the full results for all evaluated models and evaluation sets, including validity and similarity metrics.

The base Qwen3.5-4B performs poorly on this task. 
On our test set, only $27.33\%$ of its outputs are valid SMILES and only $12.53\%$ exactly match the targets, while using about $20$k output tokens on average.
After SFT using outputs from stronger LLMs, exact-match rate increases to $59.84\%$, while the average output tokens decreases to about $14$k. 
This result already exceeds several larger LLMs and approaches frontier models such as Qwen3.8-Max, GLM-5.2, and DeepSeek-V4-Pro.
RL further improves exact-match rate to $84.99\%$. 
Despite its much smaller size, the post-RL model outperforms all evaluated models except GPT-5.5 and GPT-5.6 Sol.
RL also greatly reduces the length of the reasoning by enforcing the overlong output penalty: 
the average output length decreases to only $\num{1264}$ tokens, corresponding to about a $10\times$ reduction compared with the non-GPT baselines.
Its token usage is comparable to GPT-5.6 models, still far more efficient considering the model size.


The trained model also generalizes well to the in-the-wild MolLangBench sets, though exact-match rates are lower than on MolLangData. 
On the core and extended sets, exact-match rate improves from $12.5\%$/$17.0\%$ before training to $47.0\%$/$57.0\%$ after SFT, and further to $65.0\%$/$72.0\%$ after RL. 
The RL-trained model shows similar reductions in output tokens on both MolLangBench sets.




Additionally, high exact-match rates of frontier models such as GPT-5.6 Sol and substantial gains from training a small model directly on MolLangData, further support the quality of our descriptions.

\section{Additional Use Cases, Challenges, and Future Perspectives}

Although frontier LLMs have much improved on structure-grounded tasks, 
the way they interact with molecular structures differs fundamentally from that of chemists.
LLMs rely on extensive knowledge of SMILES syntax and chemical nomenclature, 
and require substantial reasoning over these rules to process the structure.
Even the most token-efficient models for structure generation, our post-RL model and GPT-5.6 Sol, still use around $1$k tokens, with most spent enumerating atoms and applying rules to construct SMILES, as shown in Appendix Fig.~\ref{fig:reasoning_example}.
Chemists, by contrast, typically recognize structures and draw them directly, 
without explicitly reasoning through the linear notation.
Molecule--language modeling should therefore aim not only for higher accuracy, 
but also for more efficient and cognitively natural ways to perceive and generate molecular structures.

A more elegant direction is to couple language models with specialized graph-based generators having only tens of millions of parameters~\citep{defog}. 
However, long and fine-grained descriptions (Appendix~\ref{sec:appendix_dataset_statistics})
make direct generation of complete molecular graphs challenging.
LLMs can condense the descriptions without compromising completeness and fidelity, 
which Appendix~\ref{sec:description_compaction} shows is feasible on our dataset,  methodologically and economically. 
Condensation alone is  only a mitigation.
A potential solution is to introduce intermediate subgraph supervision, instead of training only on the final  molecular graph.
One could use an LLM to generate such step-by-step subgraph construction traces from our descriptions, with the target molecule used to verify the final graph.

For structure perception, once an automatic verifier, such as the introduced generation model, is reliable, 
a molecule--language model could similarly be trained with RL.
For tasks requiring only structural information, a practical compromise is to generate descriptions on demand using our annotation pipeline, without cross-modal training.
Preliminary experiments in Appendix~\ref{sec:downstream_demonstration} show that these descriptions help LLMs understand molecular structures and improve downstream property prediction.
The descriptions also provide complementary structural information for text-based molecule retrieval, 
as demonstrated by a simple retrieval system in Appendix~\ref{sec:retrieval_system}.

While MolLangData reflects the chemical space of its sample pool, PubChem, it covers common fragments and connectivity patterns, and our pipeline can scale beyond it.
The dataset and cross-modal training provide a foundation for aligning language with molecules, or, more broadly, graphs.

\subsection*{AI use statement}

In this work, we used generative AI tools for synthetic data generation, initial data validation, and partial code implementation.
We used generative AI to polish the language of the \emph{final} manuscript, limited to linguistic aspects only.
We did not use LLMs to organize the manuscript or draft any sentences from scratch.
The research ideas and method design were developed entirely by the authors.
We did not use generative AI to propose the central scientific contributions of this work.
We describe the scope of AI use in MolLangData curation and language-conditional molecule generation below.

For MolLangData curation and validation:
\begin{itemize}[leftmargin=*, 
	]
	\item We used LLMs to generate the molecular structure descriptions in MolLangData, as reported in the paper.
	\item We used LLMs in the initial stage of dataset validation, as reported in the paper.
	\item We wrote the core rule-based logic for extending OPSIN and constructing enriched XML metadata case by case, with limited AI assistance, and reviewed this code line by line.
	We used AI coding agents, specifically ChatGPT Codex, to assist with peripheral code related to compilation, testing, and code organization;
	the authors tested this code for functional correctness.
\end{itemize}

For the language-conditional molecule generation experiments:
\begin{itemize}[leftmargin=*,  
	]        
	\item We evaluated 16 LLMs as baselines on the generation benchmarks, as reported in the paper.
	\item We used LLM-generated outputs to construct the supervised fine-tuning data, as reported in the paper.
	\item The training code was developed with substantial assistance from AI coding agents, specifically ChatGPT Codex, following detailed instructions from the authors.
	The authors conducted extensive, repeated testing and debugging, provided feedback to the agents, and verified that the code behaved as intended.
\end{itemize}

We take full responsibility for the final content of this work, including text, claims, and artifacts produced with the aid of generative AI.

\subsection*{Ethics statement}

This work studies the alignment of molecular structures and natural language. 
The chemicals in our curated dataset are sourced from publicly available datasets, and the curated descriptions only faithfully describe the structure of the chemicals. 
Within this scope, the authors declare that no specific ethical concerns need to be reported.

\subsection*{Reproducibility Statement}
\label{sec:reproducibility}
To support reproducibility of the core contributions of this work,
we provide code, datasets, and model checkpoints in public repositories.
We list the artifacts and their host sites below for MolLangData
and for our language-conditional molecule generation experiments,
which we call LangMolDiode.

For MolLangData curation and validation:
\begin{itemize}[leftmargin=*]
	\item The annotation pipeline is available at \url{https://github.com/TheLuoFengLab/MolLangData}. 
	It includes source code for dataset preprocessing and LLM job submission to generate structure descriptions, along with a compiled executable for enriched XML metadata construction. 
	\item The Java source code for the structural metadata generation tool is available at \url{https://github.com/feiyang-cai/opsin_mollangdata}.
	\item The complete dataset, including approximately $161$k training samples and $2$k validation samples along with their validation results, is available at \url{https://huggingface.co/datasets/ChemFM/MolLangData}.
\end{itemize}

For the LangMolDiode experiments:
\begin{itemize}[leftmargin=*]
	\item The source code for training Qwen3.5-4B, including both SFT and RL, is available at \url{https://github.com/feiyang-cai/LangMolDiode}.
	\item The SFT training corpus, including reasoning traces and final outputs, is available at \url{https://huggingface.co/datasets/ChemFM/LangMolDiode-SFT-Corpus}.
	\item The adapter checkpoints after SFT and RL are available at \url{https://huggingface.co/collections/ChemFM/langmoldiode-checkpoints}.
\end{itemize}

Due to the inherent stochasticity of LLM API calls, the generated descriptions and validation results may not be reproduced exactly.
Our validation on $2$k samples provides an estimate of description precision, and we expect regenerated descriptions to achieve similar precision.
Likewise, we expect repeated language-conditional molecule generation evaluations, for both the baseline LLMs and our trained models, to fall within a reasonable range of the results reported in this work.


\subsubsection*{Acknowledgments}
This work was supported as part of the AIM for Composites, an Energy Frontier Research Center funded by the U.S. Department of Energy, Office of Science, Basic Energy Sciences at Clemson University under award \#DE-SC0023389.

We thank Dr.~Yongkai Wu of Clemson University for generously providing access to Azure AI Foundry to support our dataset curation and benchmark evaluations.
We also acknowledge support from the Gemini Academic Program Award, which provided credits for access to the Gemini API.
We thank the RCD LLM Service at Clemson University for providing API access to open-weight large language models hosted on Clemson's research computing infrastructure to support our dataset curation and benchmark evaluations.

\bibliography{iclr2027_conference}
\bibliographystyle{iclr2027_conference}

\newpage
\appendix

%

\begingroup
\renewcommand{\contentsname}{Technical Appendices}
\etocsetnexttocdepth{subsection}
\addcontentsline{toc}{part}{Appendix}
\localtableofcontents
\endgroup

\counterwithin{table}{section}
\counterwithin{figure}{section}
\setlength{\intextsep}{12pt plus 2pt minus 2pt}
\newpage
\section{Related Works}
\label{sec:related_works}

Molecule--language cross-modal research predates the LLM era. 
Pilot works such as Text2Mol~\citep{text2mol} and MolT5~\citep{molt5} curate paired language descriptions and molecules from public chemical databases such as PubChem~\citep{pubchem} and ChEBI~\citep{chebi}, 
where the descriptions primarily cover physicochemical and pharmacological properties with only coarse structural information. 
These datasets are then used to train models to translate such descriptions into molecular structures.

With the emergence of LLMs, their remarkable capabilities have motivated applications across diverse scientific disciplines, including chemistry.
Researchers have sought to leverage the flexibility, reasoning capacity, and extensive knowledge embedded in LLMs to tackle general chemistry problems, 
as evaluated by benchmarks such as ChemBench~\citep{chembench} and Humanity's Last Exam~\citep{lastexam}.
LLMs have also been explored for a range of molecular tasks long studied in the machine learning community, including property prediction~\citep{moleculenet}, synthesis prediction and planning~\citep{usptofull}, reaction yield prediction~\citep{suzuki_miyaur,buchwald_hartwig}, molecular optimization~\citep{mol_opt}, and \textit{de novo} molecular generation~\citep{guacamol,moses}.
These tasks and their associated datasets have been rephrased into natural-language form, with efforts including benchmarking LLM performance~\citep{chemllmbench} and curating instruction-tuning datasets~\citep{mol_instruct}.
This has led to the development of molecule--language multi-modal models,
including general-purpose models targeting multiple chemical tasks and chemical question answering~\citep{instructmol,chemllm,chemdfm,hight}, and task-specific models for property prediction~\citep{gimlet}, molecular design~\citep{llamole}, and retrosynthesis prediction~\citep{retrodfm}.

Such models have yet to demonstrate compelling performance on par with traditional uni-modal chemical models~\citep{chemfm}. 
We argue that the underlying cause is the gap in molecular structure--language alignment. 
As benchmarked by MolLangBench~\citep{mollangbench}, even state-of-the-art LLMs remain far from perfect on intuitive tasks such as molecular structure recognition and manipulation. 
Since chemical reasoning always begins with accurate structure understanding for chemists, so should AI models. 
Without reliable structure--language alignment, the powerful reasoning capacity and rich knowledge of LLMs cannot be effectively exploited, nor can downstream performance be expected to improve.

Encouragingly, several recent works have begun to recognize that structure--language alignment is central to molecule--language cross-modal modeling. 
MolTextNet~\citep{moltextnet} extracts structural features using RDKit~\citep{rdkit}, relying on $90$ functional groups defined by SMARTS patterns, but this captures only the presence of an incomplete set of functional groups, losing important structural information such as stereochemistry, ring topology, and substituent positioning. 
Mol-LLaMA~\citep{molllama} prompts GPT-4o with IUPAC names to generate structure-level molecular descriptions. 
Similarly, KnowMol~\citep{knowmol} combines both ideas, first identifying $82$ common functional groups using RDKit and then feeding the detected groups together with SMILES and IUPAC names into GPT-4o to generate structural descriptions. 
Besides the incompleteness of these descriptions, the fundamental issue is reliability: as both we and MolLangBench~\citep{mollangbench} demonstrate, even models considerably stronger than GPT-4o still fail to consistently produce accurate structural descriptions, as confirmed by manual inspection of their released datasets. 
In contrast, this work uses a rule-based approach to guide reliable description generation, producing structurally unambiguous descriptions that preserve complete structural details, such that the full molecule can be reconstructed from the description alone.

\newpage
\section{Additional Information for Dataset Curation and Validation}
\subsection{Additional Information for the Illustrative Example in Figure~\ref{fig:main_example}}
\label{appdix:running_example}
\begin{figure}[!h]
	\centering
\begin{tikzpicture}[
	font=\scriptsize,
	img/.style      ={inner sep=0pt},
	molbox/.style  ={draw=gray, densely dashed, rounded corners, fill=cyan!5,
		inner xsep=1pt, inner ysep=6pt},   
	molbox2/.style  ={draw=gray, densely dashed, rounded corners, fill=cyan!5,
		inner sep=0.5pt},   
	question/.style ={draw, rounded corners, fill=blue!5,
		align=left, text width=\linewidth-12pt, inner sep=6pt},
	answer/.style   ={draw, rounded corners, fill=cyan!5,
		align=left, text width=4.6cm, inner sep=16pt}
	]
	
	
	\node[molbox, anchor=north east, align = left, ] (mol) at (\linewidth, 0) {
		\scriptsize\bfseries \ Molecular Structure:\\[0pt]
		\includegraphics[width=4.2cm]{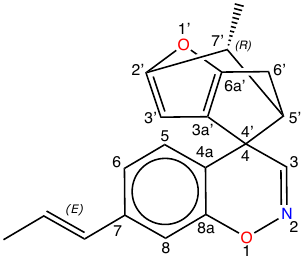}};
	
	\coordinate (mol_upper_right)    at (mol.north east);
	\coordinate (mol_lower_right)    at (mol.south east);
	\coordinate (mol_lower_left)    at (mol.south west);
	
	\coordinate (description_anchor1)    at ([xshift=-4pt, yshift=-4pt]mol.south west);
	\coordinate (description_anchor2)    at ([xshift=0pt, yshift=-4pt]mol.south east);
	\coordinate (description_anchor3)    at ([xshift=-4pt]mol.north west);
	\coordinate (rightedge) at ($(0,0) + (\linewidth,0)$);
	\coordinate (description_anchor4) at (rightedge |- description_anchor3);
	\coordinate (description_anchor5) at (\linewidth, -8.3);
	\coordinate (description_anchor6) at (0, -8.3);
	\coordinate (description_anchor7) at ($(0,0)$);	
	
	\draw[rounded corners, fill=blue!5]
	(description_anchor3) -- (description_anchor1) -- (description_anchor2) -- (description_anchor5) -- (description_anchor6) -- (description_anchor7) -- cycle;

	\path let
	\p1 = (description_anchor7),
	\p2 = (description_anchor1)
	in
	node[
	anchor=north west,
	align=left,
	text width={\x2-\x1-10pt},
	inner sep=5pt
	]
	at (description_anchor7)
	{
		\textbf{Generated Description:} \\
		\vspace{4pt}	
		Build a spiro compound in which one tetrahedral carbon is shared by two different ring systems (a "spiro" junction).
		
		\vspace{4pt}	
		
		Unprimed ring system (benzo-fused 1,2-oxazine).  Define a fused bicyclic framework and label its atoms 1, 2, 3, 4, 4a, 5, 6, 7, 8, and 8a as follows.  Make a six-member 1,2-oxazine ring containing adjacent heteroatoms: atom 1 is an oxygen (O1) and atom 2 is a nitrogen (N2).  Connect O1–N2 by a single bond, then make N2 double-bonded to carbon atom 3 (C3).  Bond C3 by a single bond to carbon atom 4 (C4), and bond C4 by a single bond to carbon atom 4a (C4a).  Close the oxazine ring by connecting C4a to carbon atom 8a (C8a) (this C4a–C8a bond is the fusion bond to a benzene ring) and then connect C8a back to O1 by a single bond.
		
		\vspace{4pt}	
		
		Fuse a benzene ring onto the oxazine by using C4a and C8a as the two shared adjacent ring atoms: the benzene ring is the six-member aromatic ring C4a–C5–C6–C7–C8–C8a (with C4a–C8a being the shared edge).  At C7, attach a prop-1-en-1-yl substituent: the
	};
	
\node[
anchor=north west,
align=left,
text width=\linewidth-10pt,
inner sep=5pt]
at (description_anchor7|-description_anchor2)
{
 ring carbon C7 is bonded to an sp2 carbon (call it P1) that is double-bonded to a second alkene carbon (P2), and P2 is single-bonded to a terminal methyl carbon (P3).  Specify the alkene geometry as E, meaning the higher-priority substituents on P1 and P2 (the ring at P1 and the methyl P3 at P2) lie on opposite sides of the P1=P2 double bond.

\vspace{4pt}	

Primed ring system (a [2,5]-methano cyclopenta[b]furan).  Label this ring system with primes as 1', 2', 3', 3a', 4', 5', 6', 6a', and 7'.  First make a five-member furan-type ring: O1'–C2'=C3'–C3a'=C6a'–O1', with double bonds specifically between C2' and C3', and between C3a' and C6a'.  Next, fuse a five-member carbocycle onto that furan by sharing the C3a'–C6a' edge: complete the carbocycle as C3a'–C4'–C5'–C6'–C6a' (all single bonds along this path, and the C3a'=C6a' bond being the shared fused edge).  The ``5',6'-dihydro" part is reflected by keeping the C5'–C6' bond as a single (saturated) bond.  Add the ``methano" bridge carbon C7' as a single carbon that is bonded to both C2' and C5' (so C2'–C7'–C5' is the bridge).  Put a methyl group on C7', and state that C7' is a stereocenter of absolute configuration R.

\vspace{4pt}	

Spiro connection.  Make the unprimed atom C4 and the primed atom C4' the same physical carbon atom (the spiro atom).  Thus this spiro carbon is bonded to C3 and C4a in the oxazine/benzene system, and to C3a' and C5' in the methanocyclopenta[b]furan system.
};

\end{tikzpicture}
\caption{Molecular structure and generated  structural description for the illustrative example shown in Fig.~\ref{fig:main_example}.
The structured metadata provided as input to the LLM are illustrated in Fig.~\ref{fig:main_example}.}
\label{fig:running_example_descriptions}

\end{figure}

\begin{figure}[h]
	\centering
\begin{tikzpicture}[
	font=\scriptsize,
	img/.style      ={inner sep=0pt},
	molbox/.style  ={draw=gray, densely dashed, rounded corners, fill=cyan!5,
		inner xsep=1pt, inner ysep=6pt},   
	molbox2/.style  ={draw=gray, densely dashed, rounded corners, fill=cyan!5,
		inner sep=0.5pt},   
	question/.style ={draw, rounded corners, fill=blue!5,
		align=left, text width=12cm, inner sep=6pt},
	answer/.style   ={draw, rounded corners, fill=cyan!5,
		align=left, text width=5.6cm, inner sep=16pt}
	]

\node[draw=none, dashed, rounded corners, font=\normalsize](title){Native OPSIN-Generated XML Parse Tree for the Illustrative Example:};

\node[draw=none, dashed, rounded corners, anchor=north west, fill=gray!10] at ([yshift=-0.3em] title.south west)
{
	\begin{minipage}{0.985\linewidth}
		\begin{minted}[fontsize=\scriptsize, style=manni
			,tabsize=1, breaklines,
			breakanywhere]{xml}
<molecule name="(7'R)-7'-methyl-7-((E)-prop-1-en-1-yl)-5',6'-dihydrospiro[benzo[e][1,2]oxazine-4,4'-[2,5]methanocyclopenta[b]furan]">
	<wordRule wordRule="simple" type="full" value="(7'R)-7'-methyl-7-((E)-prop-1-en-1-yl)-5',6'-dihydrospiro[benzo[e][1,2]ox...">
		<word type="full" value="(7'R)-7'-methyl-7-((E)-prop-1-en-1-yl)-5',6'-dihydrospiro[benzo[e][1,2]ox...">
			<substituent locant="7'">
				<group type="chain" subType="alkaneStem" value="C" labels="numeric">meth</group>
				<suffix type="inline" value="yl">yl</suffix>
				<hyphen value="-">-</hyphen>
			</substituent>
			<bracket locant="7">
				<substituent>
					<group type="chain" subType="alkaneStem" value="CCC" labels="numeric">prop</group>
					<suffix type="inline" value="yl" locant="1">yl</suffix>
				</substituent>
				<hyphen value="-">-</hyphen>
			</bracket>
			<root>
				<group type="ring" subType="ring" value="pent[b]furan" labels="1/2/3/4/5">spirobenzo[e]oxazinpent[b]furan</group>
			</root>
		</word>
	</wordRule>
</molecule>
		\end{minted}
	\end{minipage}
};
\end{tikzpicture}
\caption{Molecular structure XML parse tree produced by the native OPSIN tool~\citep{opsin} after the structure assembly for the illustrative example shown in Fig.\ref{fig:main_example}. The corresponding XML representation generated by our approach is shown in Fig.\ref{fig:main_example} for comparison.}
\label{fig:running_example_opsin_xml}
\end{figure}

\subsection{Dataset Prefiltering Statistics}
\label{appendix:filter_statistics}

We randomly sample $\num{200000}$ molecules from PubChem~\citep{pubchem} and apply prefiltering before description generation.
The breakdown is as follows:
\begin{enumerate}[label=(\roman*), leftmargin=*, labelindent=0pt, align=left, topsep=0pt, itemsep=0pt]
	\item $\num{28908}$ multi-component entries (containing `.' in SMILES) are excluded to keep validation focused on individual molecular structures.
	OPSIN and our pipeline natively support these entries, and each component can also be processed separately.
	
	\item $\num{1705}$ entries lack IUPAC names in PubChem.
	
	\item $\num{765}$ entries have mismatches between the PubChem SMILES and the structure reconstructed by OPSIN from the IUPAC name.
	Most are due to inconsistencies in the PubChem records rather than errors in our pipeline.
	For example, the PubChem SMILES for CID~136130191 specifies two E/Z configurations, but its IUPAC name specifies only one.
	
	\item $\num{1206}$ entries produce OPSIN errors or warnings. Some are due to issues in the PubChem IUPAC names, such as ambiguous locants, rather than limitations of OPSIN or our pipeline.
	
	\item The remaining $\num{167416}$ molecules are sent to LLMs for description generation. After atom-matching filtering and excluding incomplete API calls, we obtain $\num{163111}$ final descriptions.
\end{enumerate}

Most exclusions are therefore due to our focus on single-component molecules or issues in the PubChem records.
Even an overestimate that counts all cases in categories (iii) and (iv) as pipeline failures gives a failure rate of only $0.99\%$ among the $\num{200000}$ candidates, which is acceptable given the complexity of IUPAC nomenclature.

The dataset inevitably reflects the chemical space sampled from PubChem rather than the full chemical space.
However, molecular structure descriptions focus on fragments, substituents, and connectivity, and our dataset includes diverse examples of these features and connection topologies.
\subsection{Generation Difficulty Categories}
\label{appendix:difficulty_category}

We classify molecules as easy, medium, or hard based on ring topology 
and automatically route them to different LLM reasoning effort levels to balance description quality and generation cost.
\begin{itemize}[leftmargin=0.8em, labelsep=0.5em, itemsep=1pt, parsep=0pt, topsep=-2pt]
	\item \textbf{Easy:} No fused ring systems; only isolated rings and/or acyclic chains, including rings sharing a single spiro atom.
	\item \textbf{Medium:} Exactly one fused ring system containing two rings, with no spiro or bridged junctions within that system.
	\item \textbf{Hard:} A fused ring system containing more than two rings, multiple fused ring systems, or fused systems with spiro or bridged connectivity.
\end{itemize}
These categories indicate generation difficulty but do not necessarily reflect overall molecular complexity. 
Even molecules in the easy category can be structurally complex, as illustrated in Sec.~\ref{sec:data_samples}.


\subsection{Ablation Study}
\label{sec:ablation_study}
\label{sec:appendix_ablation}
We perform ablation studies to analyze the impact of major design choices in our generation pipeline.

\textbf{Effect of Structure Metadata}
We ablate structural metadata to evaluate whether injecting this information improves description precision.
Specifically, we regenerate descriptions for the same set of molecules used in the atom-matching filtering study, using identical generation models but providing only the SMILES and IUPAC name, without structural metadata (see Appendix Prompt~\ref{prompt:desp_gen_wo_metadata}).

Removing structural metadata reduces sampling efficiency, as reflected by the fraction of molecules passing the atom-matching check dropping
from {$97.7\%$} to {$93.6\%$}.
%
We then apply the same hybrid validation procedure to samples that pass atom matching (Appendix Table~\ref{tab:ablation-metadata-atomfilter}).
Although descriptions generated without metadata achieve a relatively high precision of {$94.1\%$}, this  is approximately {$4.5$} percentage points below the baseline with metadata. 
%
The performance gap increases at higher difficulty levels.
Structural metadata explicitly provides ring fusion, bridged and spiro relationships, and ring-system labeling information that otherwise requires interpretation of IUPAC names or SMILES, helping the model generate more accurate and unambiguous descriptions.
During validation, we observed that descriptions generated without structural metadata often omit fused-ring information and labeling schemes. 
Such descriptions are not considered incorrect in validation, since expert validators can still unambiguously reconstruct the exact structure using their knowledge of chemical nomenclature.
However, they are less suitable for structure--language alignment.
In contrast, validators confirm that descriptions generated with metadata consistently include these details.
The benefit of structural metadata goes beyond the improvement in precision shown in the table: it also makes structural details more explicit in the descriptions.

	\textbf{Effect of Atom-Matching Filtering}
To evaluate this effect, we validate $\num{2000}$ samples with the same difficulty-level distribution as the main validation set.
%
This set largely overlaps with the main validation set reported in Table~\ref{tab:data_generation_stat}: the main validation set is constructed by taking the passing samples from this ablation set and supplementing with additional passing samples to reach $\num{2000}$.
Among these samples, {$97.7\%$} pass the atom-matching check, indicating that the filter preserves the vast majority of generated data and does not significantly reduce generation efficiency.

We then apply the same hybrid validation procedure described above to both the passing and failing subsets.
For samples that pass the atom-matching filter, description precision remains comparable to the main validation results, achieving an overall precision of $98.6\%$.
In contrast, precision drops sharply to $27.7\%$ for the small subset that fails the filter, 
demonstrating that atom-matching effectively removes incorrectly labeled samples.
Detailed results are reported in Table~\ref{tab:ablation-metadata-atomfilter}.

We should note that the non-hydrogen atom count is requested together with description generation in the same LLM call (Prompt~\ref{prompt:description_generation}).
Since the prompt also provides SMILES as a reference for description generation, this introduces a potential risk of information leakage into the atom count.
We explicitly instruct the model to count atoms \emph{only from the generated description and not from the SMILES}.
This lightweight check helps detect missing atoms in long side chains and linkers, 
as supported by the results above.
Requesting the count separately, using only the generated description, would make the check independent of the reference SMILES but add substantial cost at our dataset scale.
Our shared-prompt design therefore balances filtering effectiveness and generation cost, while description quality is assessed separately through our strict validation protocol.

 	 \begin{table}[H]
 	 		\vspace{-0.18in}
	\centering
	\caption{Validation precision under structural metadata and atom-matching ablations across generation difficulty levels.
		Each entry is reported as passed / total evaluated samples (precision $\%$).}
	\label{tab:ablation-metadata-atomfilter}
	\small
	\begin{tabular}{@{}l
			>{\centering\arraybackslash}p{0.25\linewidth}
			>{\centering\arraybackslash}p{0.25\linewidth}
			>{\centering\arraybackslash}p{0.28\linewidth}@{}}
		\toprule
		& \multicolumn{2}{c}{With structural metadata} & \multirow{2}{*}{\shortstack{Without structural metadata\\(atom match)}} \\
		\cmidrule(lr){2-3}
		& Atom match & Atom mismatch &  \\
		\midrule
		Easy   & $\num{1275}/\num{1292}$ $(98.7\%)$ & $8/25$ $(32.0\%)$ & $\num{1191}/\num{1242}$ $(95.9\%)$ \\
		\midrule
		Medium & $472/476$ $(99.2\%)$& $5/20$ $(25.0\%)$ & $427/457$ $(93.4\%)$ \\
		\midrule
		Hard   & $178/185$ $(96.2\%)$& $0/2$ $(0.00\%)$ & $142/172$ $(82.6\%)$ \\
		\midrule
		\textbf{Overall}  & $\num{1925}/\num{1953}$ $(98.6\%)$ & $13/47$ $(27.7\%)$ & $\num{1761}/\num{1871}$ $(94.1\%)$ \\
		\bottomrule
	\end{tabular}

\end{table}

\begin{wraptable}{r}{0.50\linewidth}
	\vspace{-0.18in}
	\caption{Ablation study on model sensitivity. Pass rates ($\%$) of the LLM-based validator for descriptions generated by GPT-5 and GPT-5.2 across different difficulty levels.}
	\label{tab:gpt5-comparison}
	\small
	\begin{tabular}{lccc}
		\toprule
		Generation Model & Easy & Medium & Hard \\
		\midrule
		GPT-5   & $88.9\%$ & $66.0\%$ & $42.5\%$ \\
		GPT-5.2 & $89.5\%$ & $84.0\%$ & $63.2\%$ \\
		\bottomrule
	\end{tabular}
\end{wraptable}
\textbf{Sensitivity to Generation Model} Description generation requires LLMs to have strong chemical knowledge and the ability to reason over long, information-dense contexts.
To study model sensitivity, we replace GPT-5.2 with GPT-5~\citep{openai5} for description generation on the same molecule set used in prior ablations.
To reduce human validation cost, 
we use GPT-5.2 as the sole validator for both models and evaluate descriptions that pass the atom-matching check.
The validation pass rate for GPT-5-generated descriptions is comparable to that of GPT-5.2 in the easy category ($0.6$ percentage points lower), but is $18.0$ and $20.7$ percentage points lower in the medium and hard categories, respectively.
Detailed results are reported in  Table~\ref{tab:gpt5-comparison}.
Because we use only LLM validation, which can produce false negatives, the actual precision gap may be smaller.
These results suggest that description reliability is affected by generation model capacity, particularly for complex structures. 
Other commercial LLMs may perform differently depending on their chemical reasoning capabilities. 
Our choice of LLMs balances generation quality and budget constraints at the time of dataset curation.


\subsection{Model Snapshots}
\label{appdix:model_snapshots}
We use multiple LLMs for dataset collection and validation, 
as well as baseline comparisons in the language-conditional molecule generation experiments.
We access GPT models through Azure AI Foundry APIs, and \texttt{claude-fable-5}~\citep{claudefable5}, \texttt{gemini-3.7-flash}~\citep{gemini3_7}, and \texttt{qwen3.8-max}~\citep{qwen3_8} through their official platforms.
The remaining models are open-weight models accessed through the RCD LLM Service at Clemson University.
Table~\ref{tab:model_benchmark_configuration} lists the model snapshots or checkpoint commit hashes, as applicable.

\begin{table}[!h]
	\centering
	\small
	\caption{Model snapshots for data generation, validation, and the conditional generation benchmarks in Table~\ref{tab:model_benchmark_results}. When the snapshot date is unavailable, we report the release date. For open-weight models, we report the commit hash and date.}
	\label{tab:model_benchmark_configuration}
	\begin{tabular}{ll}
		\toprule
		Model & Snapshot \\
		\midrule
		\multicolumn{2}{l}{\textit{Qwen models}} \\
		\texttt{qwen3.5-4b} & {2026-03-01 (commit\_hash: \href{https://huggingface.co/Qwen/Qwen3.5-4B/commit/851bf6e806efd8d0a36b00ddf55e13ccb7b8cd0a}{851bf6e806efd8d0a36b00ddf55e13ccb7b8cd0a})}  \\
		\texttt{qwen3.6-35b-a3b-fp8} & {2026-04-23 (commit\_hash: \href{https://huggingface.co/Qwen/Qwen3.6-35B-A3B-FP8/commit/95a723d08a9490559dae23d0cff1d9466213d989}{95a723d08a9490559dae23d0cff1d9466213d989})}  \\
		\texttt{qwen3.8-27b-fp8} & {2026-08-14 (commit\_hash: \href{https://huggingface.co/Qwen/Qwen3.8-27B-FP8/commit/017b9c7af6b5689d5dd426a76e0bc077eb5ca20a}{017b9c7af6b5689d5dd426a76e0bc077eb5ca20a})} \\
		\texttt{qwen3.8-max} & {2026-08-02 (Release date)} \\
		\midrule
		\multicolumn{2}{l}{\textit{GLM models}} \\
		\texttt{glm-5.1-fp8} & {2026-04-07 (commit\_hash: \href{https://huggingface.co/zai-org/GLM-5.1-FP8/commit/28d85cc22ceeee52340e6ec3399bda31852b117c}{28d85cc22ceeee52340e6ec3399bda31852b117c})} \\
		\texttt{glm-5.2-fp8} & {2026-06-17 (commit\_hash: \href{https://huggingface.co/zai-org/GLM-5.2-FP8/commit/a0b55e88465d1a06afece97bc8d6b366aff39089}{a0b55e88465d1a06afece97bc8d6b366aff39089})} \\
		\midrule
		\multicolumn{2}{l}{\textit{DeepSeek models}} \\
		\texttt{deepseek-v4-pro} & {2026-08-13} \\
				\midrule
		\multicolumn{2}{l}{\textit{Google models}} \\
		\texttt{gemma-4-31b} & {2026-04-02 (commit\_hash: \href{https://huggingface.co/google/gemma-4-31B/commit/5dc307ed18be972e945cea1bfd3facf5bfd5fa39}{5dc307ed18be972e945cea1bfd3facf5bfd5fa39})} \\
		\texttt{gemini-3.7-flash} & {2026-08-13 (Release date)} \\
				\midrule
		\multicolumn{2}{l}{\textit{Claude models}} \\
		\texttt{claude-fable-5} & {2026-06-09 (Release date)} \\
		\midrule
		\multicolumn{2}{l}{\textit{GPT models}} \\
		\texttt{gpt-5} & {2025-08-07} \\
		\texttt{gpt-5.2} & {2025-12-11} \\
		\texttt{gpt-5.4} & {2026-03-05} \\
		\texttt{gpt-5.5} & {2026-04-23} \\
		\texttt{gpt-5.6-luna} & {2026-07-09 (Release date)} \\
		\texttt{gpt-5.6-terra} & {2026-07-09 (Release date)} \\
		\texttt{gpt-5.6-sol} & {2026-07-09 (Release date)} \\
		\bottomrule
	\end{tabular}
\end{table}

\subsection{Validation Statistics}
\label{sec:validation_stat}
We report validation statistics for a subset of $\num{2000}$ samples across easy, medium, and hard difficulty levels.
These samples are separate from the molecules used to develop and refine the XML pipeline.
Each LLM reconstruction attempt uses a separate call without conversation history.
Human validators work independently, without sharing reconstructions or feedback.
Both LLM and human reconstruction use the same exact-match criterion: RDKit parses and sanitizes the reconstructed and ground-truth SMILES, converts them to canonical isomeric SMILES, and compares the resulting strings.
This check preserves encoded stereochemistry, formal charges, and isotope information.

Among $\num{1317}$/$\num{496}$/$\num{187}$ samples in the easy/medium/hard categories,
GPT-5.2 validates $\num{1181}$/$\num{415}$/$\num{117}$ samples, 
GLM-5.1 validates $\num{1050}$/$\num{312}$/$\num{65}$ samples, 
and DeepSeek-V4-Pro validates $\num{1177}$/$\num{415}$/$\num{113}$ samples.
Of these, $\num{950}$/$\num{262}$/$\num{44}$ samples pass all three LLM validators and are accepted as correct, leaving $\num{744}$ samples for human validation.

Human validators assess whether each description unambiguously specifies a unique molecular structure.
Using only the description, they reconstruct the molecule in chemical drawing software and submit its SMILES to our internal validation application for the exact-match check.
A sample is approved only if the expert confirms that the description is unambiguous and the reconstructed molecule exactly matches the ground truth.

The first human validator approves $\num{343}$/$\num{230}$/$\num{136}$ samples,
and the remaining cases are forwarded to a second validator.
The second validator approves only $7$ additional easy samples and no additional medium or hard samples.
The final counts are therefore $\num{1300}$/$\num{492}$/$\num{180}$, totaling $\num{1972}$ correctly described samples.


The validation time for a single human annotator is $\qty{13.5}{\minute}$/$\qty{8.3}{\minute}$/$\qty{12.2}{\minute}$ for the easy, medium, and hard categories, respectively, with an overall average of $\qty{11.7}{\minute}$ across all validated samples.
This is slightly longer than the approximately $\qty{10}{\minute}$ reported in MolLangBench~\citep{mollangbench}.
We note that these times are collected automatically by our internal validation application and are intended as a coarse reference rather than a precise measure of annotation efficiency.



\subsection{Empirical Evaluation of False-Positive Rate in LLM Reconstruction}
\label{sec:false_positive_validation}

To reduce human validation effort, 
we combine LLM-based and human validation, as described in Secs.~\ref{sec:data_validation} and~\ref{sec:validation_stat}.
This protocol assumes that LLMs are unlikely to reconstruct the exact target molecule from an incorrect description by chance, so we expect a near-zero false-positive rate.
Exact reconstruction requires correctly recovering every structural detail from descriptions averaging more than $250$ words for molecules with approximately $27$ heavy atoms.
These details include chemical groups, connectivity, ring topology, labeling semantics, and stereochemistry, as illustrated by the representative examples in Sec.~\ref{sec:data_samples}.
Even a small error can change the resulting structure, making an exact match from an incorrect description almost impossible.

Even for the $\num{626}$ samples accepted by one or two LLMs, but not \emph{all} three, and thus rejected by automatic validation, human validation confirms unambiguous descriptions and exact reconstruction of the target molecules.
Of these, $\num{451}$ are accepted by exactly two models: $\num{304}$ are rejected only by GLM-5.1, $\num{87}$ only by GPT-5.2, and $\num{60}$ only by DeepSeek-V4-Pro.
The remaining $\num{175}$ are accepted by only one model: $\num{93}$ by GPT-5.2, $\num{24}$ by GLM-5.1, and $\num{58}$ by DeepSeek-V4-Pro.


We require successful reconstruction by \emph{all} three LLMs to provide an additional safeguard.
To evaluate this criterion, 
we randomly select $200$ samples that pass all three LLM validators for additional human validation.
Following the standard protocol, human validators receive only the textual descriptions and reconstruct the molecules from them.
All $200$ descriptions are confirmed to be unambiguous, and the corresponding molecules are reconstructed exactly, with no false positives observed.
These results provide empirical support for the reliability of LLM-based validation step.

\subsection{Dataset Statistics}
\label{sec:appendix_dataset_statistics}

\begin{figure}[!h]
	\centering
	\includegraphics[width=\linewidth]{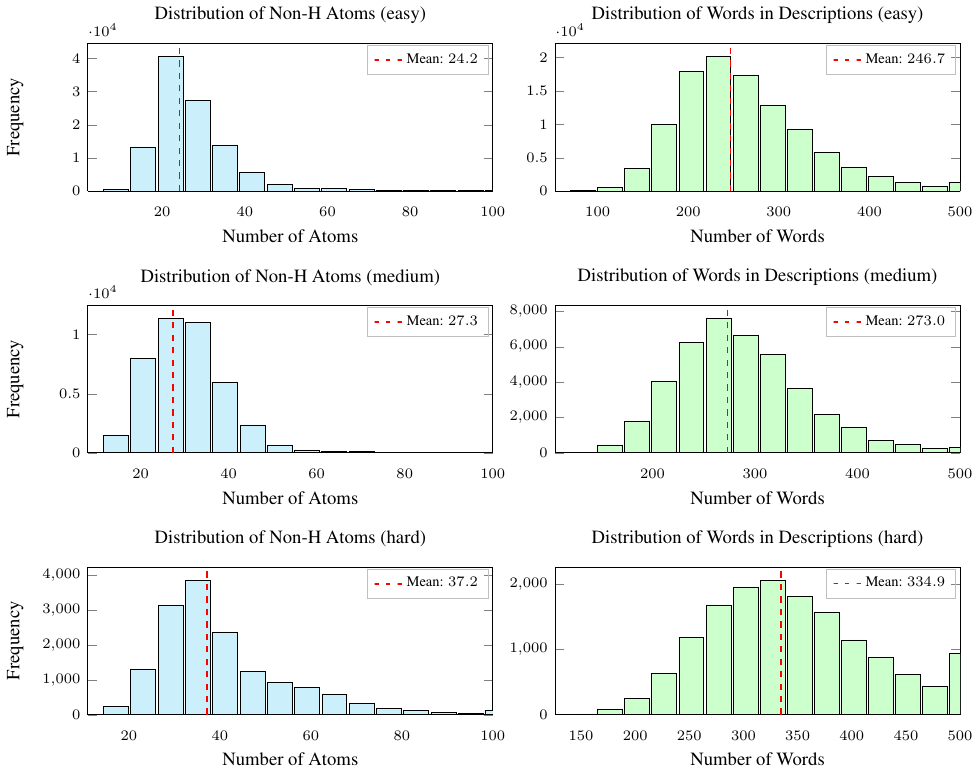}
	\caption{Distribution statistics of the entire generated dataset, including non-hydrogen atom counts and description word counts, across easy, medium, and hard categories.}
	\label{fig:data_distribution}
\end{figure}

We summarize the distribution statistics of the entire generated dataset across easy, medium, and hard difficulty levels in Fig.~\ref{fig:data_distribution}.
For the easy category, molecules span from $2$ to $335$ non-hydrogen atoms (mean $24.2$), with description lengths ranging from $53$ to $1308$ words (mean $246.7$).
The medium category covers molecules with $8$ to $274$ atoms (mean $27.3$) and descriptions of $105$ to $981$ words (mean $273.0$).
The hard category exhibits the highest structural and linguistic complexity, with atom counts ranging from $11$ to $414$ (mean $37.2$) and description lengths from $126$ to $1358$ words (mean $334.9$).
%

\subsection{Token Usage and Cost Estimation}
\label{appendix:data_generation_cost}

Generating a large-scale dataset of molecular structure descriptions incurs substantial LLM token usage and associated API costs. 
We report token and cost statistics estimated from a representative subset of $\num{2000}$ sample generations.
For this subset, the average token usage per generation is $\num{3845.35}$ input tokens and 
$\num{17089.12}$ output tokens (including $\num{16555.43}$ reasoning tokens).

As per the \href{https://platform.openai.com/docs/pricing}{OpenAI API Pricing} as of \mbox{2026-04-28} (\text{\$}$0.875$ per $1$M input tokens and \text{\$}$7.00$ per $1$M output tokens under batch mode), the estimated cost per generation is \text{\$}$0.123$. 
Scaling to the full curated dataset of $163$k samples, the total estimated cost is more than \text{\$}$20$k.

%

\subsection{Feasibility of Description Condensation}
\label{sec:description_compaction}

Our dataset contains structurally complex molecules, and we impose no verbosity constraints during description generation. 
The resulting descriptions are therefore often lengthy, as reported in Appendix~\ref{sec:appendix_dataset_statistics}.
To demonstrate the feasibility of automated condensation, we use GPT-5 with medium reasoning effort to condense descriptions for the same $200$ samples used in the false-positive evaluation in Appendix~\ref{sec:false_positive_validation}.
This reduces the average description length from $249$ to $129$ words, a $48\%$ reduction.
Human validators successfully reconstruct all $200$ molecules solely from the condensed descriptions, showing that condensation preserves reconstruction fidelity in these samples.
Condensation uses an average of $687$ input tokens and $\num{1757}$ reasoning and output tokens per sample. At this usage level, condensing the full $163$k-sample dataset would cost approximately \$$\num{1475}$.


\clearpage
\section{Additional Information for Language-Conditional Molecule Generation}
\label{sec:appdix_rl_training}

\subsection{Supervised Fine-tuning}
\label{appdix:sft}
As discussed in Sec.~\ref{sec:language_conditioned_generation}, we first perform supervised fine-tuning (SFT) because the base Qwen3.5-4B model performs poorly on this task.
We use outputs from more capable LLMs, including their reasoning traces, to improve the initial policy and support faster convergence during subsequent RL training.
We use the $\num{161111}$ MolLangData training samples, which are separate from the $\num{1972}$ test samples, to collect molecule-generation outputs from GLM-5.1 and Qwen3.6-35B.
Samples are routed to these models according to the availability of our institution's hosted LLM services; 
not every sample is sent to both models.
After deduplication, we obtain $\num{80850}$ outputs that exactly match the target molecules: $\num{62286}$ from GLM-5.1 and $\num{18564}$ from Qwen3.6-35B.
We discard outputs longer than $\num{40960}$ tokens, leaving $\num{75666}$ samples for SFT, comprising $\num{62185}$ from GLM-5.1 and $\num{13481}$ from Qwen3.6-35B.
Both the reasoning traces and final answers are used for training.

We fine-tune Qwen3.5-4B using an $85$M-parameter LoRA adapter for $3$ epochs, with a batch size of $16$ samples per optimization step.
Training uses AdamW with a learning rate of $\num{2e-4}$ and a weight decay of $0.01$, following a cosine learning-rate schedule with $20$ warm-up steps.
SFT takes approximately $\qty{19.1}{\hour}$ on a single DGX H100 node with $8$ H100 GPUs, each with $\qty{80}{\giga\byte}$ of memory.
SFT improves the exact-match accuracy of molecule generation on the MolLangData test set from $12.53\%$ to $59.84\%$, providing a stronger starting point for subsequent RL training.

\subsection{Reinforcement Learning}
\label{appdix:rl}

After SFT, 
we use DAPO~\citep{dapo} on the same $\num{161111}$ training samples to further improve molecule-generation accuracy and reasoning efficiency.
\begin{itemize}[leftmargin=-0pt,  
	label={},        
	 topsep=0pt,
	itemsep=0pt,
	parsep=\parskip,
	wide=0pt]        
\item[\textbf{Reward Design}] 
Let $\mathcal{M}$ be the target molecule and $\widehat{\mathcal{M}}$ the molecule parsed from a generated response.
We penalize responses with missing or invalid SMILES.
For valid predictions, we assign rewards based on structural similarity and exact matching to the target.
The base reward $R_{\mathrm{base}}$ is
\begin{equation*}
	R_{\mathrm{base}} =
	\begin{cases}
		-1.0, & \text{if no SMILES can be extracted},\\
		-0.4, & \text{if the extracted SMILES is invalid},\\
		\operatorname{sim}(\mathcal{M},\widehat{\mathcal{M}})
		+0.4\times\mathbbm{1}(\mathcal{M}=\widehat{\mathcal{M}})+0.1,
		& \text{otherwise}.
	\end{cases}
	\label{eq:rl-base-reward}
\end{equation*}
Here, $\operatorname{sim}$ is the Tanimoto similarity between Morgan fingerprints of the generated and target.
We compute fingerprints with $2048$ bits and radius $2$.
$\mathbbm{1}(\mathcal{M}=\widehat{\mathcal{M}})$ evaluates whether their canonical isomeric SMILES match, using the same exact-match criterion as in our dataset validation.

To discourage long reasoning, we apply a linear penalty beyond $L_\text{min}=\num{2048}$ response tokens, reaching $-0.4$ at the maximum length of $L_\text{max}=\num{32768}$ tokens:
\begin{equation*}
	R_{\mathrm{len}}(L)
	=
	-0.4\times\operatorname{clip}\!\left(
	\frac{L-L_\text{min}}{L_\text{max}-L_\text{min}},0,1
	\right),
	\label{eq:rl-length-penalty}
\end{equation*}
where $L$ is the response length in tokens.
The final reward is
\begin{equation*}
	R_{\mathrm{final}}=R_{\mathrm{base}}+R_{\mathrm{len}}(L).
	\label{eq:rl-final-reward}
\end{equation*}

\item[\textbf{Training Configuration}] 
We implement DAPO using verl~\citep{verl}, with  vLLM engines~\citep{vllm} for rollout  and the Megatron backend~\citep{megatron} for distributed LoRA policy optimization.
Each rollout step targets $512$ prompt groups, with eight responses per prompt.
We compute advantages within each group and use DAPO's token-level clipped policy objective, without a KL-loss term.
In practice, all eight responses in a group may receive identical rewards, 
so their advantages are zero and they provide no policy-gradient signal.
Following DAPO's dynamic sampling strategy, we sample $640$ prompts, discard groups with identical rewards, and sample additional batches as needed to collect $512$ informative groups.
The retained groups are divided into $16$ mini-batches, each containing $32$ prompts and their $256$ responses.
Each mini-batch is used for one optimization step, giving $16$ optimization steps per rollout.

With this sampling procedure, we continue training the LoRA adapter from SFT for $\num{1500}$ rollout steps, totaling $\num{24000}$ optimizer updates.
The learning rate increases linearly from zero to $10^{-6}$ over the first $20$ rollout steps and remains constant thereafter.
Training takes approximately $\qty{322.6}{\hour}$ on two DGX H100 nodes with $16$ NVIDIA H100 GPUs in total.

\end{itemize}

\subsection{Benchmark Results}
\label{sec:model_benchmark}

We benchmark  language-conditional molecule generation on the MolLangData test set ($\num{1972}$ samples held out from training) and the MolLangBench core and extended sets ($200$ samples each)~\citep{mollangbench}.
We check all three test sets against the training set using the same exact-match criterion based on canonical isomeric SMILES and confirm  there is no overlap.

We evaluate Qwen3.5-4B before training, after SFT, and after RL.
We compare these results with $16$ much larger LLMs, including Qwen, GLM, DeepSeek, Gemini, Claude, and GPT models.
Their snapshots or release dates are listed in Table~\ref{tab:model_benchmark_configuration}.
We extract SMILES from model responses and report validity, along with Tanimoto similarity (invalid outputs scored as $0$) and exact-match rate relative to the ground truth.
We also report the average number of output tokens to compare token efficiency.
Full results are provided in Table~\ref{tab:model_benchmark_results}.
For reference, we also report the best exact-match rate achieved during RL, independently for each test set.


All baseline models use medium reasoning effort except GLM-5.1 and Qwen3.6-35B, which enable reasoning without selectable effort levels, and GLM-5.2, which uses high.
Because of frequent API refusals, Claude Fable 5 results use only non-refusal responses and are for reference: $\num{1750}$ of $\num{1972}$ samples on MolLangData, 
$196$ of $200$ on MolLangBench core and all $200$ on the extended set.
%

\begin{table*}[h]

\centering
\scriptsize
\setlength{\tabcolsep}{0.9pt}
\caption{Language-conditional molecule generation on MolLangData and MolLangBench. Validity and exact-match rate (Exact) are percentages; average output tokens are rounded to integers.}
\label{tab:model_benchmark_results}
\begin{tabular}{p{1.05in}cccccccccccc}
\toprule
\multirow{2}{0.9in}{\centering Model} & \multicolumn{4}{c}{MolLangData (Test; 1,972 Samp.)} & \multicolumn{4}{c}{MolLangBench (Core; 200 Samp.)} & \multicolumn{4}{c}{MolLangBench (Extended; 200 Samp.)} \\
\cmidrule(lr){2-5}\cmidrule(lr){6-9}\cmidrule(lr){10-13}
& Validity & Similarity & Exact & Tokens & Validity & Similarity & Exact & Tokens & Validity & Similarity & Exact & Tokens \\
\midrule
\multicolumn{13}{l}{\textit{Qwen models}} \\
\texttt{qwen3.6-35b} & 71.04\% & 0.6317 & 46.42\% & 51,451 & 75.00\% & 0.6379 & 43.50\% & 41,832 & 75.50\% & 0.6495 & 47.00\% & 46,793 \\
\makecell[l]{\texttt{qwen3.8-27b}} & 76.52\% & 0.6748 & 42.75\% & 8,915 & 76.50\% & 0.6717 & 41.00\% & 8,956 & 85.50\% & 0.7556 & 53.50\% & 5,918 \\
\makecell[l]{\texttt{qwen3.8-max}} & 86.97\% & 0.8167 & 63.54\% & 6,707 & 82.00\% & 0.7500 & 53.00\% & 6,776 & 91.50\% & 0.8478 & 69.00\% & 4,578 \\
\midrule
\multicolumn{13}{l}{\textit{GLM models}} \\
\texttt{glm-5.1-fp8} & 70.91\% & 0.6391 & 45.57\% & 13,899 & 73.00\% & 0.6401 & 43.50\% & 17,091 & 80.00\% & 0.7090 & 53.50\% & 11,485 \\
\texttt{glm-5.2-fp8} & 88.87\% & 0.8351 & 65.18\% & 10,912 & 86.50\% & 0.7798 & 53.00\% & 12,652 & 91.00\% & 0.8373 & 68.00\% & 7,936 \\
\midrule
\multicolumn{13}{l}{\textit{DeepSeek models}} \\
\texttt{deepseek-v4-pro} & 87.69\% & 0.8451 & 68.21\% & 18,469 & 89.50\% & 0.8589 & 66.00\% & 22,907 & 93.50\% & 0.8988 & 76.00\% & 13,335 \\
\midrule
\multicolumn{13}{l}{\textit{Google models}} \\
\texttt{gemma-4-31b} & 58.11\% & 0.4535 & 25.91\% & 12,251 & 54.00\% & 0.3801 & 17.00\% & 11,159 & 69.00\% & 0.5409 & 30.50\% & 10,095 \\
\texttt{gemini-3.7-flash} & 97.82\% & 0.9571 & 81.44\% & 2,527 & 98.00\% & 0.9419 & 75.00\% & 2,926 & 99.00\% & 0.9667 & 82.50\% & 2,035 \\
\midrule
\multicolumn{13}{l}{\textit{Claude models (partial results; reference only)}} \\
\texttt{claude-fable-5} & 99.20\% & 0.9831 & 92.74\% & 627 & 98.98\% & 0.9823 & 92.35\% & 898 & 99.50\% & 0.9606 & 86.50\% & 550 \\
\midrule
\multicolumn{13}{l}{\textit{GPT models}} \\
\texttt{gpt-5} & 72.46\% & 0.6490 & 49.39\% & 7,539 & 66.00\% & 0.5585 & 38.50\% & 8,044 & 76.00\% & 0.6886 & 56.00\% & 6,164 \\
\texttt{gpt-5.2} & 87.73\% & 0.8314 & 68.56\% & 3,700 & 89.50\% & 0.8299 & 64.00\% & 4,554 & 93.00\% & 0.8649 & 71.00\% & 2,514 \\
\texttt{gpt-5.4} & 94.42\% & 0.9184 & 81.95\% & 3,495 & 94.50\% & 0.9016 & 74.00\% & 4,274 & 95.50\% & 0.9337 & 84.00\% & 2,528 \\
\texttt{gpt-5.5} & 98.38\% & 0.9756 & 92.95\% & 2,623 & 99.00\% & 0.9783 & 93.50\% & 2,949 & 100.00\% & 0.9779 & 93.50\% & 1,827 \\
\texttt{gpt-5.6-luna} & 93.51\% & 0.9061 & 78.09\% & 1,625 & 92.50\% & 0.8857 & 69.50\% & 1,868 & 95.00\% & 0.9114 & 80.00\% & 1,242 \\
\texttt{gpt-5.6-terra} & 92.65\% & 0.8810 & 73.48\% & 1,219 & 95.00\% & 0.8777 & 67.50\% & 1,328 & 94.50\% & 0.8980 & 76.50\% & 843 \\
\texttt{gpt-5.6-sol} & 99.65\% & 0.9929 & 94.27\% & 1,058 & 99.00\% & 0.9775 & 88.50\% & 1,172 & 99.50\% & 0.9824 & 95.00\% & 730 \\
\midrule
\multicolumn{13}{l}{\textit{Qwen3.5-4B training stages}} \\
\makecell[l]{pre-SFT; base} & 27.33\% & 0.2136 & 12.53\% & 19,338 & 23.50\% & 0.1985 & 12.50\% & 22,241 & 28.00\% & 0.2490 & 17.00\% & 20,058 \\
\makecell[l]{post-SFT} & 84.33\% & 0.7874 & 59.84\% & 14,304 & 78.00\% & 0.6924 & 47.00\% & 16,362 & 87.50\% & 0.7847 & 57.00\% & 13,162 \\
\makecell[l]{post-RL} & 97.41\% & 0.9581 & 84.99\% & 1,264 & 93.50\% & 0.8700 & 65.00\% & 1,430 & 97.50\% & 0.9155 & 72.00\% & 1,067 \\
\makecell[l]{best-checkpoint} & 96.91\% & 0.9551 & 85.19\% & 1,277 & 93.50\% & 0.8681 & 71.50\% & 1,273 & 97.00\% & 0.9248 & 77.50\% & 1,294 \\
\bottomrule
\end{tabular}
\vspace{-0.2in}
\end{table*}

\subsection{Example of Molecule Generation Reasoning}

Fig.~\ref{fig:reasoning_example} shows an example output from our post-RL Qwen3.5-4B model.
Although the molecule contains only a two-ring fused system with three substituents, the model uses $590$ output tokens to generate its SMILES representation.
Much of this output is used to enumerate atom positions and apply SMILES syntax rules to construct and repeatedly check the resulting string.

\begin{figure}[H]
		\centering
		\input{./figures/reasoning_example/example.tex}
		\caption{An example output from our post-RL Qwen3.5-4B model.}
		\label{fig:reasoning_example}
\end{figure}

\section{Additional Experiments for Dataset Usage}
\label{sec:data_usage}
Beyond molecule–language cross-modal training, we briefly explore a few additional, preliminary use cases enabled by our annotation pipeline and curated dataset.

\subsection{Downstream Task Evaluation of Description Generation Pipeline}
\label{sec:downstream_demonstration}
Our annotation pipeline generates accurate and complete molecular structure descriptions, 
which can be directly used for tasks requiring structural information as input. 
We demonstrate that the automatically generated structural descriptions can benefit 
(i) molecular structure recognition, evaluated on the MolLangBench~\citep{mollangbench} 
benchmark, and (ii) downstream chemical property reasoning, evaluated on the 
MoleculeNet~\citep{moleculenet} benchmark. 
For both benchmarks, molecular structures are 
first converted to IUPAC names using the OpenEye LexiChem Toolkit~\citep{lexichem}, which 
are then passed to our pipeline for structural description generation.

\begin{itemize}[leftmargin=-0pt,  
	label={},        
	topsep=0pt,
	itemsep=0pt,
	parsep=\parskip,
	wide=0pt]        

\item[\textbf{Molecular structure recognition.}] 
We select $10$ molecular recognition subtasks from MolLangBench that do not require 
identifying specific target atoms (such as identifying $k$-hop neighbors, determining 
bond connectivity, or characterizing the specified atom or bond stereochemistry), as such 
tasks would require manually specifying atoms in natural language when using descriptions 
as input. 
Localization accuracy is also excluded for the same reason. 
For each query, we generate a structural description using our pipeline and prompt GPT-5, the best-performing model evaluated in MolLangBench, at medium reasoning effort to perform the recognition task solely based on the description.

The results are reported in Table~\ref{tab:mollangbench_recognition_gpt5_comparison_selected}. 
Despite SMILES representations already achieving strong performance at $0.963$ average 
accuracy, our generated structural descriptions match or improve performance on $9$ out of 
$10$ subtasks ($3$ retain perfect scores of $1.000$ and $6$ improve), reaching $0.985$ 
average accuracy. 
Notably, ring junction recognition (i.e., atoms 
shared between two or more rings) improves by $0.105$, suggesting that structural 
descriptions effectively assist LLMs in understanding molecular topology. 
These results demonstrate that descriptions generated by our pipeline can enhance structural understanding 
in LLM-centric AI systems.
		
\begin{table}[H]
	\centering
	\small
	\caption{Molecular structure recognition results on MolLangBench~\citep{mollangbench}, 
		comparing SMILES representations against structural descriptions generated by our 
		pipeline. Both are evaluated using GPT-5 with medium reasoning effort; the SMILES 
		baseline results are taken directly from MolLangBench.}
	\begin{tabular}{lccc}
		\toprule
		Task & SMILES & Structural Description & $\Delta$ \\
		\hline
		Aldehyde              & $1.000$ & $1.000$ &  $0.000$ \\
		Benzene               & $0.925$ & $0.995$ & $+0.070$ \\
		Carboxyl              & $0.980$ & $0.985$ & $+0.005$ \\
		Furan                 & $0.995$ & $1.000$ & $+0.005$ \\
		Halogen atoms         & $1.000$ & $1.000$ &  $0.000$ \\
		Ketone                & $1.000$ & $1.000$ &  $0.000$ \\
		Pyridine              & $0.915$ & $0.955$ & $+0.040$ \\
		Quaternary carbons    & $0.980$ & $1.000$ & $+0.020$ \\
		Ring junctions        & $0.830$ & $0.935$ & $+0.105$ \\
		Thiophene             & $1.000$ & $0.985$ & $-0.015$ \\
		\midrule
		Average               & $0.963$ & $0.985$ & $+0.022$ \\
		\bottomrule
	\end{tabular}
	\label{tab:mollangbench_recognition_gpt5_comparison_selected}
\end{table}
	
	\item[\textbf{Molecular property prediction.}] 
To demonstrate that improved structural understanding further enhances downstream chemical 
reasoning, we evaluate on two tasks from the MoleculeNet benchmark: BACE, a binary 
classification task for inhibitors of human $\beta$-secretase 1, and ESOL, a regression 
task for water solubility with continuous labels in $\log\,\unit{\mol/\liter}$. 
We compare  two molecular representations as input: SMILES and our generated structural descriptions, 
both evaluated with GPT-5 using medium reasoning effort. 
Note that GPT-5 is a reasoning model that internally analyzes molecular structure even when given only SMILES as input.

Following MoleculeNet, we use scaffold splits for BACE and random splits for ESOL.
Results are reported in Table~\ref{tab:bace-esol-summary-gpt5} as mean $\pm$ standard 
deviation across three splits. 
Our structural descriptions improve performance by $5.03\%$ 
on BACE and $11.97\%$ on ESOL, directly demonstrating that our pipeline enhances LLM-based 
chemical reasoning. 
We expect the advantage to be even larger in tasks requiring fine-grained 
structural information, such as forward/retro-synthesis prediction or molecular property 
optimization.
	
\end{itemize}

\begin{table}[H]
	\centering
	\small
	\caption{Molecular property prediction results on the BACE and ESOL datasets from MoleculeNet~\citep{moleculenet}, comparing SMILES representations against structural descriptions generated by our pipeline. Both are evaluated using GPT-5 with medium reasoning effort. Values are mean $\pm$ standard deviation across three splits, with bold indicating the better performance.}
	\label{tab:bace-esol-summary-gpt5}
	\begin{tabular}{lcc}
		\toprule
		Dataset & SMILES & Structural Description \\
		\midrule
		BACE (Accuracy $\uparrow$) & $0.656 \pm 0.014$ & $\mathbf{0.689 \pm 0.016}$ \\
		ESOL (RMSE $\downarrow$) & $0.844 \pm 0.024$ & $\mathbf{0.743 \pm 0.073}$ \\
		\bottomrule
	\end{tabular}
\end{table}

\subsection{Preliminary Demonstration of Text-Based Molecular Retrieval}


\label{sec:retrieval_system}

\begin{figure}[h]
	
	\centering
	
	
	\definecolor{samplebg}{RGB}{245,245,245}
	\begin{tikzpicture}
		
		
		\node[
		draw=gray,
		densely dashed,
		rounded corners,
		fill=cyan!5,
		inner xsep=4pt, inner ysep=4pt,
		anchor=north
		] (fig1) at (0,0)
		{\includegraphics[width=\textwidth]{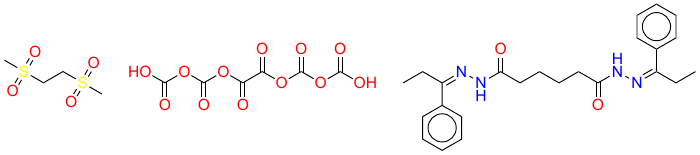}};
		
		\node[below=0.15cm of fig1.south, align=center]
		(txt1)
		{Text query 1: \textit{``a molecule with a symmetric structure''}};
		
		
		\node[        draw=gray,
		densely dashed,
		rounded corners,
		fill=cyan!5,
		inner xsep=4pt, inner ysep=4pt,
		anchor=north,
		below=0.4cm of txt1.south] (fig2)
		{\includegraphics[width=\textwidth]{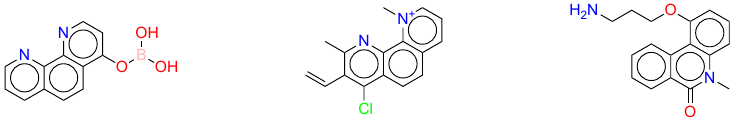}};
		
		\node[below=0.15cm of fig2.south, align=center]
		(txt2)
		{Text query 2: \textit{``an angular fused tricyclic aromatic system''}};
		
		
		\node[        draw=gray,
		densely dashed,
		rounded corners,
		fill=cyan!5,
		inner xsep=4pt, inner ysep=4pt,
		anchor=north,
		below=0.4cm of txt2.south] (fig3)
		{\includegraphics[width=\textwidth]{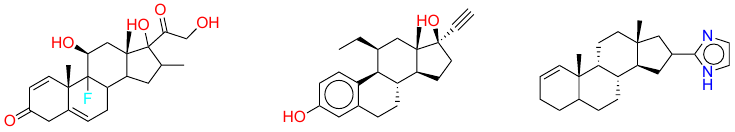}};
		
		\node[below=0.15cm of fig3.south, align=center]
		{Text query 3: \textit{``a molecule with a steroid-like fused ring system''}};
		
	\end{tikzpicture}
	
	\caption{Text-based molecular retrieval over our 
		$163$k generated descriptions. Each row shows the top-3 retrieved molecules for the corresponding natural-language structural query shown below.}
	
	\label{fig:retrieval_systems}
	
\end{figure}

Descriptions in current public chemical repositories, such as PubChem~\citep{pubchem} and ChEBI~\citep{chebi}, contain only coarse structural information and physicochemical or pharmacological properties. 
Our generated descriptions, which capture complete structural details, can serve as complementary information for textual molecule retrieval systems.

We provide a preliminary demonstration using a simple dense retrieval system over our
$163$k generated descriptions. 
The index combines full-description embeddings with chunk-level embeddings. 
At the full-description level, we precompute one embedding per molecule using text-embedding-3-large~\citep{openai_embedding}. 
At the chunk level, each description is split into overlapping 50-word chunks with 20-word overlap, and each chunk is independently embedded using the same model. 
At query time, the user query is embedded and searched against both indices. 
Chunk-level matches are aggregated back to the molecule level by taking the best chunk score for each molecule, with a small bonus when multiple chunks from the same molecule are retrieved. 
The final ranking is obtained by combining the full-description and chunk-level rankings using weighted reciprocal rank fusion.

Fig.~\ref{fig:retrieval_systems} shows the top-$3$ retrieval results for three example queries of different structural specificity. 
The retrieved molecules are consistent with the intended structural characteristics described in each query, suggesting that our generated descriptions can support meaningful text-based molecular retrieval. 
Notably, the first two queries describe structural patterns that are difficult to express as SMARTS~\citep{smarts} regular expressions. 
We emphasize that this is a preliminary demonstration; more sophisticated retrieval architectures and systematic evaluation are left for future work.

\section{Prompts and Representative Examples}

%
%

\subsection{Prompt for Language-Conditional Molecule Generation}
We use the prompt from Appendix A.21.3 of MolLangBench~\citep{mollangbench}, listed below, to ask an LLM to generate a molecule in SMILES format from a detailed language description.
This prompt is used both for LLM-based validation in Sec.~\ref{sec:data_validation} and for the language-conditional molecule generation experiments in Sec.~\ref{sec:language_conditioned_generation}.

\begin{promptbox}[prompt:validation]{Molecule Generation from Natural-Language Descriptions}
	\label{prompt:molecule_generation}
	\footnotesize
	{\normalsize \textbf{Instructions:}}
    
    \vspace{4pt}
    
	Analyze the given molecule structure description and generate the corresponding molecular structure as SMILES representation.
	
	\begin{itemize}[leftmargin=1.2em]
		\item This description provides an unambiguous and detailed account sufficient to completely reconstruct the molecule.
		\item Interpret stereochemistry, charges, and isotopes if mentioned in the description.
		\item Return the resulting SMILES string enclosed within $<$smiles$>$ and $<$/smiles$>$ tags.
	\end{itemize}

	\vspace{4pt}
	
	{\normalsize \textbf{Molecule structure description:}}
	
	\vspace{4pt}
	{\color{xmlcolor} \texttt{\{Structure Description\}}}

\end{promptbox}

\subsection{Description Generation Prompts and Complex-Ring XML Semantics}
\label{appdix:generation_prompt}
\begin{promptbox}[prompt:desp_gen]{Molecular Description Generation from SMILES, IUPAC Names, and Structural Metadata}
	\footnotesize
	\label{prompt:description_generation}
	{\normalsize \textbf{Task:}}
	
	\vspace{4pt}
	You are provided with the \textbf{IUPAC name}, \textbf{SMILES string}, and \textbf{rule-based hierarchical metadata} describing the molecular structure.
	Generate a \textbf{detailed and accurate structural description} that would allow a person with \textbf{basic organic chemistry knowledge} to reconstruct the exact molecule using only your structural description text.
	The description should be \textbf{natural}, \textbf{diverse}, and \textbf{chemically precise}, clearly conveying the molecular skeleton, substituents, and stereochemistry.
	
	\vspace{4pt}
	In addition, after completing the description, report the total number of non-hydrogen atoms implied by your description, computed \textbf{only from the information explicitly stated in the description itself}.
	
	\vspace{8pt}
	{\normalsize \textbf{Guidelines:}}
	\begin{enumerate}[leftmargin=1.2em]
		
		\item \textbf{Purpose and Independence}
		\begin{itemize}[leftmargin=1.2em]
			\item The description must be \textbf{self-contained} and \textbf{sufficient for reconstruction}.
			\item Assume the reader only has your final text---they will \textbf{not see} the SMILES, IUPAC name, or metadata.
			\item You may use all input data internally to reason about the molecule, but the final description must read as a stand-alone, human-readable explanation.
		\end{itemize}
		
		\item \textbf{Freedom of Description}
		
		You may begin from any perspective---the main skeleton, a key ring system, or an important substituent.
		Combine information freely from the IUPAC name, SMILES, and metadata to capture the complete structure.

		\item \textbf{Backbone and Connectivity}
		
		Describe how rings, chains, and substituents are connected.
		Indicate branching positions, linkages, and overall topology so that the structure can be reconstructed accurately.
		
		\item \textbf{Functional Groups and Substituents}
		
		Identify all key functional groups and substituents.
		Specify their type (e.g., hydroxyl, amine, halogen, carbonyl), location, and bonding pattern relative to the molecular framework.
		
		\item \textbf{Simple or Isolated Rings}
		
		For common rings (benzene, pyridine, cyclohexane), you may name them directly or briefly describe their composition and bonding.
		
		\item \textbf{Fused, Bridged, or Spiro Ring Systems --- explicit, structured, and verified}
		
		It is \textbf{strongly recommended to follow the guidance below} when describing complex ring topologies:
		
		When interpreting fused, bridged, or spiro ring systems, you should explicitly refer to the corresponding ring-system semantics described later. These semantics are essential for correctly understanding atom labeling, fusion points, bridge connections, and spiro locants as derived from the provided metadata.
		
		\begin{enumerate}[label=(\alph*),leftmargin=1.7em]
			\item \textbf{Define and label atoms/rings.}
			Explicitly assign ring labels and atom labels to each ring in the ring system (e.g., ``Ring A: C1–C6 clockwise starting at the junction with ring B''; include heteroatoms like O1/N5 as needed).
			For complex fused or multi-ring systems, it is recommended to align the atom labeling by the structure metadata to maintain chemical accuracy.
			If there are multiple distinct fused systems, assign separate, non-overlapping label sets to avoid confusion.
			
			\item \textbf{Describe internal ring features.}
			State ring size, aromaticity/saturation, heteroatoms, and bonding patterns (alternation, single/double).
			
			\item \textbf{Explain how rings are connected.}
			Specify \textbf{exact shared atoms or edges} (e.g., ``Ring B shares the C5–C6 edge with ring A'').
			Describe \textbf{fusion geometry} (linear, angular, bridged, spiro).
		\end{enumerate}
		
		\textbf{Required verification (before finalizing your description):}
		
		\begin{itemize}[leftmargin=1.2em]
			\item \textbf{Label consistency:} Each label you introduced is used consistently; no duplicate or skipped numbering within a ring.
			\item \textbf{Ring labeling check:} Verify that the atom labeling and numbering sequence are correct, and that the shared atoms or edges described for each fusion correspond accurately to your own ring definitions and orientations. The metadata should be treated as the \textbf{gold standard} for validating both labeling and fusion topology.
			\item \textbf{Orientation sanity-check:} The shared atoms/edge and the named orientation (linear vs angular; inner vs outer edge) match the topology. Do not accidentally mirror or rotate the fusion.
			\item \textbf{Cross-consistency check:} Ensure that substituent positions and stereochemical designations correspond correctly to the atom labels and numbering scheme you defined for each ring.
		\end{itemize}
		
		\item \textbf{Stereochemistry}
		
		Include stereochemical information such as (R/S) or (E/Z) when available, and describe how these configurations relate to surrounding atoms or bonds.
		
		\item \textbf{Rational Use of Metadata}
		
		\begin{itemize}[leftmargin=1.2em]
			\item Treat the metadata as \textbf{accurate structural evidence}, but express its meaning in \textbf{your own words}.
			\item The metadata is not shown to the reader, so do not include or reference any of its raw contents directly.
			\item Only mention atoms, labels, or locants (e.g., “C9,” “N5”) \textbf{after you have introduced them} in your own description.
			\item When metadata provides \textbf{partial or shorthand notations}, infer and expand these to their complete, chemically correct forms.
		\end{itemize}

		\item \textbf{Use of Chemical Shorthand}
		\begin{itemize}[leftmargin=1.2em]
			\item \textbf{Avoid full structural formulas} written as continuous symbolic notations.
			\item Short fragments like ``--OH,'' ``--CH3,'' or ``--NH2'' are acceptable when helpful.
		\end{itemize}
		
		\item \textbf{Balance and Readability}
		
		Aim for a \textbf{balanced level of detail}.
		
		\item \textbf{Descriptive Diversity}
		
		Use varied styles and sentence structures.
		
		\item \textbf{Do Not Include}
		\begin{itemize}[leftmargin=1.2em]
			\item The full IUPAC name, SMILES string, or XML tags verbatim.
			\item Long symbolic chemical formulas for the entire molecule.
			\item Brand names, trivial comments, or unrelated metadata.
			\item Unintroduced atom labels or locants.
		\end{itemize}
		
		\item \textbf{Non-hydrogen atom count}
		
		\begin{itemize}[leftmargin=1.2em]
			\item After completing the description, report the \textbf{total number of non-hydrogen atoms}.
			\item Do \textbf{not} use the SMILES to perform this count.
			\item Do \textbf{not} include the counting process in your description.
		\end{itemize}
	\end{enumerate}
	
	\vspace{8pt}
	
	{\normalsize \textbf{Output Format:}}
	\begin{xmloutputbox}
		<description>
		[Concise, varied, and chemically precise structural description]
		</description>
		<non_hydrogen_atom_count>
		[integer]
		</non_hydrogen_atom_count>
	\end{xmloutputbox}
	
	\vspace{4pt}
	\begin{structmetabox}
		{\normalsize\color{red!75!black}
			This section of the prompt specifies semantics for \emph{fused}, \emph{spiro}, and \emph{bridged} ring systems, which are automatically injected when required.
			Detailed prompt specifications for these semantics are provided in prompts
			~\ref{prompt:fused_ring_semantics},
			~\ref{prompt:bridged_ring_semantics}, and
			~\ref{prompt:spiro_ring_semantics},
			respectively.
		}
	\end{structmetabox}
	
	\vspace{10pt}
	{\normalsize \textbf{Inputs:}}
	\begin{itemize}[leftmargin=1.2em]
		\item \textbf{IUPAC Name:} {\color{xmlcolor} \texttt{\{IUPAC\}}}
		\item \textbf{SMILES String:} {\color{xmlcolor} \texttt{\{SMILES\}}}
		\item \textbf{Molecular Metadata (XML Hierarchy):}
	\end{itemize}
\begin{tcolorbox}[colback=gray!5!white, boxrule=0pt, left=6pt, right=3pt, top=-2pt, bottom=3pt, frame empty]
	{\ttfamily\color{xmlcolor}\{XML\_METADATA\}}
\end{tcolorbox}
\end{promptbox}
\begin{promptbox}[prompt:desp_gen_wo_metadata]{Molecular Description Generation from SMILES and IUPAC Names}
	\footnotesize
	
	{\normalsize \textbf{Task:}}
	
	\vspace{4pt}
	You are provided with the \textbf{IUPAC name} and \textbf{SMILES string},
	generate a \textbf{detailed and accurate structural description} that would allow a person with \textbf{basic organic chemistry knowledge} to reconstruct the exact molecule using only your structural description text.
	The description should be \textbf{natural}, \textbf{diverse}, and \textbf{chemically precise}, clearly conveying the molecular skeleton, substituents, and stereochemistry.
	
	\vspace{4pt}
	In addition, after completing the description, report the total number of non-hydrogen atoms implied by your description, computed \textbf{only from the information explicitly stated in the description itself}.
	
	\vspace{8pt}
	{\normalsize \textbf{Guidelines:}}
	\begin{enumerate}[leftmargin=1.2em]
		
		\item \textbf{Purpose and Independence}
		\begin{itemize}[leftmargin=1.2em]
			\item The description must be \textbf{self-contained} and \textbf{sufficient for reconstruction}.
			\item Assume the reader only has your final text---they will \textbf{not see} the SMILES or IUPAC name.
			\item You may use all input data internally to reason about the molecule, but the final description must read as a stand-alone, human-readable explanation.
		\end{itemize}
		
		\item \textbf{Freedom of Description}
		
		You may begin from any perspective---the main skeleton, a key ring system, or an important substituent.
		Combine information freely from the IUPAC name and SMILES to capture the complete structure.

		\item \textbf{Backbone and Connectivity}
		
		Describe how rings, chains, and substituents are connected.
		Indicate branching positions, linkages, and overall topology so that the structure can be reconstructed accurately.
		
		\item \textbf{Functional Groups and Substituents}
		
		Identify all key functional groups and substituents.
		Specify their type (e.g., hydroxyl, amine, halogen, carbonyl), location, and bonding pattern relative to the molecular framework.
		
		\item \textbf{Simple or Isolated Rings}
		
		For common rings (benzene, pyridine, cyclohexane), you may name them directly or briefly describe their composition and bonding.
		
		\item \textbf{Fused, Bridged, or Spiro Ring Systems --- explicit, structured, and verified}
		
		It is \textbf{strongly recommended to follow the guidance below} when describing complex ring topologies:
		
		\begin{enumerate}[label=(\alph*),leftmargin=1.7em]
			\item \textbf{Define and label atoms/rings.}
			Explicitly assign ring labels and atom labels to each ring in the ring system (e.g., ``Ring A: C1–C6 clockwise starting at the junction with ring B''; include heteroatoms like O1/N5 as needed).
			If there are multiple distinct fused systems, assign separate, non-overlapping label sets to avoid confusion.
			
			\item \textbf{Describe internal ring features.}
			State ring size, aromaticity/saturation, heteroatoms, and bonding patterns (alternation, single/double).
			
			\item \textbf{Explain how rings are connected.}
			Specify \textbf{exact shared atoms or edges} (e.g., ``Ring B shares the C5–C6 edge with ring A'').
			Describe \textbf{fusion geometry} (linear, angular, bridged, spiro).
		\end{enumerate}
		
		\textbf{Required verification (before finalizing your description):}
		
		\begin{itemize}[leftmargin=1.2em]
			\item \textbf{Label consistency:} Each label you introduced is used consistently; no duplicate or skipped numbering within a ring.
			\item \textbf{Ring labeling check:} Verify that the atom labeling and numbering sequence are correct, and that the shared atoms or edges described for each fusion correspond accurately to your own ring definitions and orientations.
			\item \textbf{Orientation sanity-check:} The shared atoms/edge and the named orientation (linear vs angular; inner vs outer edge) match the topology. Do not accidentally mirror or rotate the fusion.
			\item \textbf{Cross-consistency check:} Ensure that substituent positions and stereochemical designations correspond correctly to the atom labels and numbering scheme you defined for each ring.
		\end{itemize}
		
		\item \textbf{Stereochemistry}
		
		Include stereochemical information such as (R/S) or (E/Z) when available, and describe how these configurations relate to surrounding atoms or bonds.
		
		\item \textbf{Use of Chemical Shorthand}
		\begin{itemize}[leftmargin=1.2em]
			\item \textbf{Avoid full structural formulas} written as continuous symbolic notations.
			\item Short fragments like ``--OH,'' ``--CH3,'' or ``--NH2'' are acceptable when helpful.
		\end{itemize}
		
		\item \textbf{Balance and Readability}
		
		Aim for a \textbf{balanced level of detail}.
		
		\item \textbf{Descriptive Diversity}
		
		Use varied styles and sentence structures.
		
		\item \textbf{Do Not Include}
		\begin{itemize}[leftmargin=1.2em]
			\item The full IUPAC name or SMILES tags verbatim.
			\item Long symbolic chemical formulas for the entire molecule.
			\item Brand names, trivial comments, or unrelated metadata.
			\item Unintroduced atom labels or locants.
		\end{itemize}
		
		\item \textbf{Non-hydrogen atom count}
		
		\begin{itemize}[leftmargin=1.2em]
			\item After completing the description, report the \textbf{total number of non-hydrogen atoms}.
			\item Do \textbf{not} use the SMILES to perform this count.
			\item Do \textbf{not} include the counting process in your description.
		\end{itemize}
	\end{enumerate}
	
	\vspace{8pt}
	
	{\normalsize \textbf{Output Format:}}
	\begin{xmloutputbox}
		<description>
		[Concise, varied, and chemically precise structural description]
		</description>
		<non_hydrogen_atom_count>
		[integer]
		</non_hydrogen_atom_count>
	\end{xmloutputbox}

	\vspace{8pt}
	{\normalsize \textbf{Inputs:}}
	\begin{itemize}[leftmargin=1.2em]
	\item \textbf{IUPAC Name:} {\color{xmlcolor} \texttt{\{IUPAC\}}}
	\item \textbf{SMILES String:} {\color{xmlcolor} \texttt{\{SMILES\}}}
\end{itemize}
\end{promptbox}
\begin{promptbox}[prompt:fused_ring_semantics]{Fused Ring Semantics (XML)}
\label{prompt:fused_ring_semantics}
\footnotesize

A fused ring system is represented in the XML by \textbf{incrementally constructing new ring
	identities} from simpler rings. Each fusion step introduces a \textbf{new global labeling
	scheme}, which replaces all previous ones and is used for any subsequent fusion and
structural references.

\vspace{10pt}
{\normalsize \textbf{Core semantics}}
	\begin{enumerate}[leftmargin=1.2em]
		
\item \textbf{Local ring definition}
\begin{itemize}[leftmargin=1.2em, itemsep=0.pt]
	\item Each ring is described by a \texttt{value} attribute containing its
	\textbf{SMILES representation}; in some fused ring systems, this SMILES value may
	be omitted.
	\item A \texttt{labels} attribute assigns \textbf{atom labels} for the ring; when a SMILES
	value is present, the labels follow the \textbf{atom order in the SMILES}.
	\begin{itemize}[leftmargin=1.2em, itemsep=0pt]
		\item If \texttt{labels} is \textbf{explicitly provided}, those labels define the
		atom labeling scheme.
		\item If \texttt{labels} is \textbf{omitted} or set to \texttt{numeric}, atoms are
		implicitly labeled from 1 to $n$, where $n$ is the number of atoms
		in the ring, following SMILES order.
	\end{itemize}
	\item Atom indices \textbf{start from 1}.
\end{itemize}

\item \textbf{Fusion via \texttt{originalLabels}}
\begin{itemize}[leftmargin=1.2em, itemsep=2pt]
	\item \texttt{originalLabels} maps \textbf{each atom of the newly fused system} to atom
	indices in the component rings.
	\item Each entry corresponds to \textbf{one atom in the fused system}.
	\item Multiple indices in an entry indicate a \textbf{fusion point}.
	\item A blank position indicates that the fused-system atom does not belong to that ring.
\end{itemize}

\item \textbf{Label propagation}
\begin{itemize}[leftmargin=1.2em, itemsep=2pt]
	\item After fusion, the fused system receives a new \texttt{labels} list.
	\item This labeling scheme \textbf{replaces all previous local labels}.
	\item All subsequent structural references—including further fusions, bridged connections,
	spiro connections, substituent attachment positions, and stereochemical
	descriptors—\textbf{must reference this new labeling scheme}.
\end{itemize}
\end{enumerate}

\vspace{10pt}
{\normalsize \textbf{Worked Example: indeno[5,6-b]furan}}

\begin{xmlcodebox}
<group type="ring" subType="fusedRing" value="indeno[5,6-b]furan">
	<fusedChildRing
		type="ring"
		subType="ring"
		value="o1cccc1"
		labels="1/2/3/4/5">
	furan
	</fusedChildRing>
	<fusedChildRing
		type="ring"
		subType="fusionRing"
		value="[cH2]1ccc2ccccc12"
		labels="1/2/3/3a/4/5/6/7/7a"
		fusedRing1="[cH2]1cccc1"
		fusedRing2="c1ccccc1"
		originalLabels="(1,)/(2,)/(3,)/(4,1)/(,2)/(,3)/(,4)/(,5)/(5,6)">
	indeno
	</fusedChildRing>
	<fusedRingLabels
		labels="1/2/3/3a/4/4a/5/6/7/7a/8/8a"
		originalLabels="(1,)/(5,)/(4,)/(3,6)/(,7)/(,7a)/(,1)/(,2)/(,3)/(,3a)/(,4)/(2,5)">
	</fusedRingLabels>
</group>
\end{xmlcodebox}

\vspace{10pt}
{\normalsize \textbf{Step-by-step interpretation}}

\begin{enumerate}[label=\textbf{Step~\arabic*:}, leftmargin=*, itemsep=2pt]
\item \textbf{ Ring A (furan)}

\begin{itemize}[leftmargin=-1.5em]
\item SMILES: \texttt{o1cccc1}
\item Labels: \texttt{1/2/3/4/5}
\item Label assignment follows SMILES order:
\begin{itemize}[leftmargin=1.2em, itemsep=1pt]
	\item O $\rightarrow$ 1
	\item Carbons $\rightarrow$ 2--5
\end{itemize}
\end{itemize}

\item \textbf{Ring B (indeno)}
	
{\par
	\parshape=2
	1.1em \linewidth
	1.5em \dimexpr\linewidth-1.5em\relax
	Ring B is formed by fusing two sub-rings:
	\par}
\begin{itemize}[leftmargin=-1.5em, itemsep=1pt]
	\item \texttt{fusedRing1}: \texttt{[cH2]1cccc1} $\rightarrow$ atoms 1--5
	\item \texttt{fusedRing2}: \texttt{c1ccccc1} $\rightarrow$ atoms 1--6
\end{itemize}

\vspace{4pt}
{\par
	\parshape=2
	1.1em \linewidth
	1.5em \dimexpr\linewidth-1.5em\relax
	Fusion is defined by:
	\par}

\texttt{originalLabels = (1,)/(2,)/(3,)/(4,1)/(,2)/(,3)/(,4)/(,5)/(5,6)}

\vspace{4pt}
{\par
	\parshape=2
	1.1em \linewidth
	1.5em \dimexpr\linewidth-1.5em\relax
	Fusion points:
	\par}

\begin{itemize}[leftmargin=-1.5em, itemsep=1pt]
	\item atom 4 of \texttt{fusedRing1} $\leftrightarrow$ atom 1 of \texttt{fusedRing2}
	\item atom 5 of \texttt{fusedRing1} $\leftrightarrow$ atom 6 of \texttt{fusedRing2}
\end{itemize}

\vspace{4pt}
{\par
	\parshape=2
	1.1em \linewidth
	1.5em \dimexpr\linewidth-1.5em\relax
	After fusion, Ring B receives a \textbf{new labeling scheme}:
	\par}

	\texttt{labels = 1/2/3/3a/4/5/6/7/7a}

\vspace{4pt}
{\par
	\parshape=2
	1.1em \linewidth
	1.5em \dimexpr\linewidth-1.5em\relax
	From this point onward, \textbf{indeno is treated as a single ring}.
	\par}

\item \textbf{Fuse Ring A with Ring B}

{\par
	\parshape=2
	1.1em \linewidth
	1.5em \dimexpr\linewidth-1.5em\relax
	The final fused system (indeno[5,6-b]furan) is created by fusing:
	\par}

\begin{itemize}[leftmargin=-1.5em, itemsep=1pt]
	\item Ring A (labels \texttt{1/2/3/4/5})
	\item Ring B (labels \texttt{1/2/3/3a/4/5/6/7/7a})
\end{itemize}

\vspace{4pt}
{\par
	\parshape=2
	1.1em \linewidth
	1.5em \dimexpr\linewidth-1.5em\relax
	The final system receives a \textbf{new global labeling scheme}:
	\par}

	\texttt{labels = 1/2/3/3a/4/4a/5/6/7/7a/8/8a}

\vspace{4pt}
{\par
	\parshape=2
	1.1em \linewidth
	1.5em \dimexpr\linewidth-1.5em\relax
	Fusion is defined by:
	\par}

	\texttt{(1,)/(5,)/(4,)/(3,6)/(,7)/(,7a)/(,1)/(,2)/(,3)/(,3a)/(,4)/(2,5)
	}

\vspace{4pt}
{\par
	\parshape=2
	1.1em \linewidth
	1.5em \dimexpr\linewidth-1.5em\relax
	Fusion points:
	\par}

\begin{itemize}[leftmargin=-1.5em, itemsep=1pt]
	\item atom 3 of Ring A $\leftrightarrow$ atom 6 of Ring B
	\item atom 2 of Ring A $\leftrightarrow$ atom 5 of Ring B
\end{itemize}

\end{enumerate}
\end{promptbox}
\begin{promptbox}[prompt:bridged_ring_semantics]{Bridged Ring Semantics (XML)}
	
	\footnotesize
	A bridged ring system is represented by defining a \textbf{parent ring system} and a
	\textbf{bridge fragment} that connects two atoms of the parent ring.
	
	\vspace{4pt}
	For a bridged ring system, the XML normally provides:
	\begin{itemize}[leftmargin=1.2em,itemsep=0.2pt]
		\item A \textbf{parent ring system}, defined by its \texttt{value} (SMILES, when provided)
		and \texttt{labels}. The parent ring may itself be a fused ring system, following the
		fused-ring semantics.
		\item A \textbf{bridge fragment}, defined by its own \texttt{value} (SMILES) and
		\texttt{labels}.
		\item A \textbf{\texttt{bridgeLocants}} attribute, which specifies the \textbf{two atom labels
			of the parent ring} that are connected by the bridge.
	\end{itemize}
	
	The \texttt{bridgeLocants} are ordered and define the \textbf{direction along the bridge}.
	The labels of the bridge fragment are assigned \textbf{from the first locant to the second
		locant}, following the order of the bridge fragment’s SMILES.
	
	\vspace{10pt}
	{\normalsize\textbf{Worked Example: 4a,8a-propanoquinoline}}
	
	\begin{xmlcodebox}
<group type="ring" subType="bridgeSystem" value="propanoquinoline">
	<bridgeParent
		type="ring"
		subType="ring"
		value="n1cccc2ccccc12"
		labels="1/2/3/4/4a/5/6/7/8/8a"
		fusedRing1="n1ccccc1"
		fusedRing2="c1ccccc1"
		originalLabels="(1,)/(2,)/(3,)/(4,)/(5,1)/(,2)/(,3)/(,4)/(,5)/(6,6)">
	quinoline
	</bridgeParent>
	<bridgeChild
		type="chain"
		subType="alkaneStem"
		value="-CCC-"
		labels="11/10/9"
		usableAsAJoiner="yes"
		bridgeLocants="4a,8a">
	prop
	</bridgeChild>
</group>
	\end{xmlcodebox}
	
	\vspace{10pt}
	{\normalsize\textbf{Interpretation:}}
	\begin{itemize}[leftmargin=1.2em,itemsep=0.2pt]
		\item Parent ring labels: \texttt{1/2/3/4/4a/5/6/7/8/8a}
		\item Bridge locants: \texttt{4a,8a}
		\item Bridge labels: \texttt{11/10/9}
	\end{itemize}
	
	The bridge is incorporated directionally as: \textbf{4a -- 11 -- 10 -- 9 -- 8a}
	
	\vspace{4pt}
	The resulting structure uses a \textbf{single combined labeling scheme}, consisting of the
	parent-ring labels extended by the bridge labels. All subsequent references (including
	further fusions, substituents, and stereochemistry) must use this combined labeling scheme.
\end{promptbox}
\begin{promptbox}[prompt:spiro_ring_semantics]{Spiro Ring Semantics (XML)}
	
	\footnotesize
For a spiro ring system, the XML normally provides:

\begin{itemize}[leftmargin=1.2em, itemsep=2pt]
	\item One or more \texttt{spiroSystemComponent} entries, each describing a
	\textbf{complete ring system}.
	\begin{itemize}[leftmargin=1.2em, itemsep=1pt]
		\item Each component has a \texttt{value} (SMILES) and a \texttt{labels} attribute.
		\item If \texttt{labels} is \textbf{explicitly provided}, those labels are used; if
		\texttt{labels} is \textbf{omitted} or set to \texttt{numeric}, atoms are
		implicitly labeled from 1 to $n$, following the \textbf{SMILES atom
			order}.
		\item A component may itself be a fused ring system, following the fused-ring
		semantics.
	\end{itemize}
	\item A \texttt{spiroLocant}, which specifies the \textbf{spiro atom in each ring
		system}.
\end{itemize}

Each ring system \textbf{retains its own labeling scheme}. To distinguish atoms belonging
to different ring systems, \textbf{prime notation} ($'$, $''$, \ldots) is used.

\vspace{6pt}
The locants in \texttt{spiroLocant} are listed in the same order as the
	\texttt{spiroSystemComponent} entries and identify the single atom in each ring
	system that represents the same physical atom.

\vspace{10pt}
{\normalsize \textbf{Worked Example: spiro[cyclopentane-1,1$'$-indene]}}

\begin{xmlcodebox}
<group type="spiro system" subType="Polycyclic" value="spiro,pent,inden">
	<spiroSystemComponent 
		type="ring" 
		subType="alkaneStem" 
		value="C1CCCC1" 
		labels="numeric">
	pent
	</spiroSystemComponent>
	<spiroLocant>1,1'</spiroLocant>
	<spiroSystemComponent
		type="ring"
		subType="ring"
		value="[cH2]1ccc2ccccc12"
		labels="1/2/3/3a/4/5/6/7/7a"
		fusedRing1="[cH2]1cccc1"
		fusedRing2="c1ccccc1"
		originalLabels="(1,)/(2,)/(3,)/(4,1)/(,2)/(,3)/(,4)/(,5)/(5,6)">
	inden
	</spiroSystemComponent>
</group>
\end{xmlcodebox}

\vspace{10pt}
{\normalsize \textbf{Interpretation:}}
\begin{itemize}[leftmargin=1.2em, itemsep=1pt]
	\item \textbf{Cyclopentane} uses labels \texttt{1}--\texttt{5}; the spiro atom is \textbf{1}.
	\item \textbf{Indene} uses its own labeling scheme; the spiro atom is \textbf{1$'$}.
	\item Atoms \textbf{1} and \textbf{1$'$} refer to the \textbf{same physical atom}.
\end{itemize}

All other atoms are referenced using their \textbf{component-specific labels with primes},
consistent with the XML.
\end{promptbox}

\clearpage
\subsection{Representative Samples}

\label{sec:data_samples}
We present representative molecule–description pairs from each generation difficulty level (easy, medium, and hard), with two examples per category.
For each level, we select one molecule with a relatively simple structure and one with a more complex structure.
For the structurally simpler examples, we additionally include the structured XML metadata constructed by our method.

\begin{figure}[h]
   \vspace{-10.5pt}
	\begin{subfigure}{\linewidth}
		\centering
		\input{./figures/examples/easy/1.tex}
		\caption{Easy category example 1 (simple structure): \textit{methyl-[2-(2-prop-2-enoyloxyethylsulfanyl)ethyl]phosphinic acid}}
		\label{fig:easy simple}
	\end{subfigure}
\end{figure}

\clearpage
\begin{figure}[H]
	\ContinuedFloat
	\begin{subfigure}{\linewidth}
		\centering
		\input{./figures/examples/easy/2.tex}
			\caption{Easy category example 2 (complex structure): \textit{(2S)-2-[(3S,6S,9E,12R,15R,18S,21S,24R,27R)-18-(4-azanylbutyl)-15,24-bis(2-azanylethyl)-27-[[(3S,4R)-3,4-bis(oxidanyl)tetradecanoyl]amino]-3-[(1S)-2-chloranyl-1-oxidanyl-ethyl]-9-ethylidene-21-(2-hydroxy-2-oxoethyl)-12-[(1S)-1-oxidanylethyl]-2,5,8,11,14,17,20,23,26-nonakis(oxidanylidene)-1-oxa-4,7,10,13,16,19,22,25-octazacyclooctacos-6-yl]-2-oxidanyl-ethanoic acid}}
		\label{fig:easy complex}
	\end{subfigure}
\end{figure}

\begin{figure}[H]
	\ContinuedFloat
	\begin{subfigure}{\linewidth}
		\centering
		\input{./figures/examples/medium/1.tex}
	\caption{Medium category example 1 (simple structure): \textit{3,4-dihydro-2H-1,5-benzodioxepin-7-yl-(2-fluorophenyl)methanone}}
		\label{}
	\end{subfigure}
\end{figure}

\begin{figure}[H]
	\ContinuedFloat
	\begin{subfigure}{\linewidth}
		\centering
		\input{./figures/examples/medium/2.tex}
			\caption{Medium category example 2 (complex structure): \textit{N'-[(2R)-6-azanyl-1-phenylsulfanyl-hexan-2-yl]-4-(5-fluoranylquinolin-8-yl)-N-(3-nitrophenyl)sulfonyl-benzohydrazide}}
		\label{}
	\end{subfigure}
\end{figure}

\begin{figure}[H]
	\ContinuedFloat
	\begin{subfigure}{\linewidth}
		\centering
		\input{./figures/examples/hard/1.tex}

					\caption{Hard category example 1 (complex structure): \textit{4-[5-[4-[5-[4-[4,8-bis(4-fluoranyl-5-hexyl-thiophen-2-yl)-6-methyl-thieno[2,3-f][1]benzothiol-2-yl]-5,6-bis(2-ethylhexoxy)-2,1,3-benzothiadiazol-7-yl]thiophen-2-yl]-2,5-bis(fluoranyl)phenyl]thiophen-2-yl]-5,6-bis(2-ethylhexoxy)-7-methyl-2,1,3-benzothiadiazole}}
		\label{}
	\end{subfigure}
\end{figure}

\begin{figure}[H]
	\ContinuedFloat
	\begin{subfigure}{\linewidth}
		\centering
		\input{./figures/examples/hard/2.tex}
		\caption{Hard category example 2 (simple structure): \textit{N-(fluoren-9-ylideneamino)-2,3-dihydro-1,4-benzodioxine-3-carboxamide}}
		\label{}
	\end{subfigure}
	\caption{
		Representative examples for the generated dataset.}
\end{figure}

\end{document}